\pdfoutput=1

\documentclass[11pt]{article}

\usepackage[final]{acl}

\usepackage{times}
\usepackage{latexsym}

\usepackage[T1]{fontenc}

\usepackage[utf8]{inputenc}

\usepackage{microtype}

\usepackage{inconsolata}

\usepackage{graphicx}
\usepackage{amssymb}
\usepackage{amsmath}
\usepackage{multirow}
\usepackage{array}
\usepackage{enumitem}
\usepackage{booktabs}
\title{LAWCAT: Efficient Distillation from Quadratic to \underline{L}inear \underline{A}ttention \underline{w}ith \underline{C}onvolution \underline{a}cross \underline{T}okens for Long Context Modeling}

\author{
 \textbf{Zeyu Liu\textsuperscript{1}},
 \textbf{Souvik Kundu\textsuperscript{2}},
 \textbf{Lianghao Jiang\textsuperscript{1}},
 \textbf{Anni Li\textsuperscript{1}},
\\
 \textbf{Srikanth Ronanki\textsuperscript{3}},
 \textbf{Sravan Bodapati\textsuperscript{3}},
 \textbf{Gourav Datta\textsuperscript{4}},
 \textbf{Peter A. Beerel\textsuperscript{1}}
\\
 \textsuperscript{1}University of Southern California, USA
 \textsuperscript{2}Intel Labs, San Diego, USA \\
 \textsuperscript{3}Amazon AGI, USA,
 \textsuperscript{4}Case Western Reserve University, USA
\\
 \small{
   \textbf{Correspondence:} \href{mailto:email@domain}\{liuzeyu, pabeerel\}@usc.edu
 }
}

\begin{document}
  \maketitle
  \begin{abstract}
Although transformer architectures have achieved state-of-the-art performance across diverse domains, their quadratic computational complexity with respect to sequence length remains a significant bottleneck, particularly for latency-sensitive long-context applications.
While recent linear-complexity alternatives are increasingly powerful, effectively training them from scratch is still resource-intensive. 
To overcome these limitations, we propose LAWCAT (Linear Attention with Convolution Across Time), a novel linearization framework designed to efficiently transfer the capabilities of pre-trained transformers into a performant linear attention architecture.
LAWCAT integrates causal Conv1D layers to enhance local dependency modeling and employs normalized gated linear attention to improve generalization across varying context lengths. 
Our comprehensive evaluations demonstrate that, distilling Mistral-7B with only 1K-length sequences yields over 90\% passkey retrieval accuracy up to 22K tokens, significantly extending its effective context window. 
Similarly, Llama3.2-1B LAWCAT variant achieves competitive performance on S-NIAH 1\&2\&3 tasks (1K-8K context length) and BABILong benchmark (QA2\&QA3, 0K-16K context length), 
requiring less than 0.1\% pre-training tokens compared with pre-training models.
Furthermore, LAWCAT exhibits faster prefill speeds than FlashAttention-2 for sequences exceeding 8K tokens. 
LAWCAT thus provides an efficient pathway to high-performance, long-context linear models suitable for edge deployment, reducing reliance on extensive long-sequence training data and computational resources. Code is released at: \url{https://github.com/zeyuliu1037/LAWCAT}.
  \end{abstract}

  \section{Introduction}

The widespread adoption and significant success of transformer-based architectures have driven remarkable advancements across various natural language processing (NLP)~\citep{deepseekai2025deepseekr, grattafiori2024llama3herdmodels, qwen2.5}, reasoning ~\cite{chen2025seal}, and computer vision (CV) tasks~\citep{yao2024minicpmv, ramachandran2025ouromamba}. Transformers leverage self-attention mechanisms to effectively model long-range dependencies, achieving state-of-the-art (SOTA) performance on diverse benchmarks~\citep{open-llm-leaderboard-v2}. 
However, their quadratic computational complexity with respect to sequence length remains a critical bottleneck, limiting their applicability to latency-sensitive edge based applications that
are constrained in terms of training data and cost.

To address this limitation, modern recurrent models have emerged as efficient alternatives, approximating or reformulating attention mechanisms to achieve linear complexity. Recent studies, such as Mamba-2~\citep{mamba2}, Gated linear attention (GLA)~\citep{yang2023gated}, and Hierarchically gated linear RNN2 (HGRN2)~\citep{hgrn2}, demonstrate the feasibility of state update mechanisms that substantially reduce computational costs while maintaining competitive performance. Despite these advances, training these models from scratch often requires substantial computational resources and extensive data, hindering rapid development and experimentation.

On the other hand, knowledge distillation, a powerful technique for transferring knowledge from a large, pretrained teacher model to a smaller and more efficient student model, offers an effective strategy to alleviate these challenges. Recently, some research like MambaInLlama~\cite{junxiongdaniele2024mambainllama} and LoLCATs~\cite{zhang2025lolcats} directly distill transformer models into recurrent architectures. However, MambaInLlama still requires extensive training data ($\sim$20B tokens) and retains 50\% self-attention layers to achieve strong performance up to 38K tokens, and LoLCATs exhibits limited generalization to sequences significantly longer than those seen during distillation.

To address this problem, we introduce LAWCAT (Linear Attention with Convolution Across Time), a novel distillation framework that efficiently transfers and enhances the rich representational capabilities of pre-trained transformer models into modern linear attention models. LAWCAT is particularly well-suited for edge applications that require long-context modeling under tight computational constraints with limited training data.
LAWCAT integrates a depth-separable convolution module,
consisting of a depth-wise convolution layer followed by a linear layer (similar to MobileNet~\citep{howard2017mobilenets}, but we use causal Conv1D across tokens), 
and a gated linear attention with normalization to effectively bridge the gap between quadratic and linear attention mechanisms.
The main contributions of this paper are summarized as follows:
\begin{itemize}[noitemsep, topsep=0pt]
    \item We proposed the integration of a causal Conv1D layer to enhance the capacity of linear attention mechanisms to model local dependencies and introduce the normalization for the gated linear attention to improve generalization to long contexts.
    \item We present LAWCAT, a framework leveraging these components to efficiently distill high-performing linear attention models. Specifically, we show that our Mistral v0.1 7B LAWCAT variant achieves over 90\% accuracy on the passkey retrieval task with context lengths up to 22K tokens, significantly exceeding the original model's 8K limit. 
    In addition, our Llama3.2 1B model LAWCAT variant achieves comparable performance on more complex retrieval tasks (NIAH1-3) and reasoning tasks (QA2\&3 from BABILong benchmark) across various sequence lengths.
    \item We demonstrate the efficiency of our distillation approach, requiring significantly less data (less than 0.1\% of the original pre-training tokens, and less than 1/3 of the original sequence length) to adapt a pretrained transformer into a high-performing linear attention model. Additionally, we show that the resulting LAWCAT model exhibits faster prefill-stage processing speeds compared to FlashAttention-2 for input sequences longer than 8K tokens.
\end{itemize}

\section{Background and Related Work}
\textbf{Self-Attention} 
The majority of the powerful LLM models like Llama 3~\citep{grattafiori2024llama3herdmodels}, Mistral~\citep{jiang2023mistral7b}, and Phi 4~\citep{abdin2024phi} adopt the multi-head scaled dot-product softmax attention (SDPA). 
Given a query $\mathbf{q}_{t}\in \mathbb{R}^{d_q}$, keys $\{\mathbf{k}_{i}\}_{i=1}^{t}\in \mathbb{R}^{d_k}$, and values $\{\mathbf{v}_{i}\}_{i=1}^{t}\in \mathbb{R}^{d_v}$ ($t\leq N$), a single-head SDPA computes output 
$\mathbf{o}_{t}$ as (omitting the scaling factor $\sqrt{d}$ for simplicity):
\begin{equation}
   \mathbf{o}_{t}= \sum_{i=1}^{t}\frac{\exp\!\Big(\mathbf{q}_{t} \mathbf{k}_{i}^{T}\Big)}{\sum_{j=1}
    ^{t}\exp\!\Big(\mathbf{q}_t \mathbf{k}_j^T\Big)}\,\mathbf{v}_{i} 
    \label{eq:sa:itr}
\end{equation}
This can be equivalently expressed as $\mathbf{o}_{t} = \sum_{i=1}^{t} w_{ti} \mathbf{v}_{i}$, where $w_{ti}$ denotes the attention weight assigned to value $\mathbf{v}_{i}$ at output position $t$.
The softmax in the numerator/denominator ensures context-dependent normalization – each weight $w_{ti}$ depends not only on the query–key similarity $\mathbf{q}_{t}\mathbf{k}_{i}^{T}$ but also on all other similarities $\{\mathbf{q}_{t}\mathbf{k}_{j}^{T}\}_{j \le t}$. This coupling gives softmax attention a powerful ability to focus on a few relevant tokens while normalizing out less relevant ones, but also yields the $O(N^{2})$ computation complexity with respect to the sequence length $N$.


\noindent
\textbf{Kernel-Based Linear Attention} To mitigate this quadratic computation complexity, many linear attention methods aim to find a kernel function whose structure approximates attention scores through inner products of transformed query and key representations.
Assuming the kernel $\kappa$ is positive semi-definite (PSD), then there exists a feature mapping $\phi$ satisfying:
$\kappa(\mathbf{q}, \mathbf{k})= \phi(\mathbf{q})\phi(\mathbf{k})^T$
Thus, the attention output at position $t$ can be computed as:
\begin{equation}
    \begin{aligned}
      \mathbf{o}_{t} & = \sum_{i=1}^{t}{\frac{\phi(\mathbf{q}_{t}) \phi(\mathbf{k}_{i})^{T}}{\sum_{j=1}^{t}\phi(\mathbf{q}_{t}) \phi(\mathbf{k}_{j})^{T}}}\,\mathbf{v}_{i} 
                     \\& ={\frac{\phi(\mathbf{q}_{t}) \sum_{i=1}^{t}  \phi(\mathbf{k}_{i})^{T}\mathbf{v}_{i}}{\phi(\mathbf{q}_{t})\sum_{j=1}^{t} \phi(\mathbf{k}_{j})^{T}}}
    \end{aligned}
    \label{eq:la:1}
\end{equation}
\noindent
Similar to the Eq.~\ref{eq:sa:itr}, we can simplify is as $\mathbf{o}_{t} = \sum_{i=1}^{t} \tilde{w}_{ti} \mathbf{v}_{i}$, and $\tilde{w}_{ti}$ can be regarded as the attention weights of the linear attention.
This is analogous to softmax attention but uses $\phi (\mathbf{q})\phi(\mathbf{k})^{T}$ to approxiamte
$\exp(\mathbf{q}\mathbf{k}^{T})$ for all queries/keys, at least in expectation. 
If  $\phi$ can perfectly realize $\exp$ as an inner product in a higher-dimensional feature space, then the linear attention exactly equals softmax attention. 
In practice, $\phi$ might be a simple nonlinear function such as $\mathrm{1+ELU}$~\citep{katharopoulos2020transformers}, or a positive random feature map as in the Performer~\citep{choromanski2020rethinking} that aims to fit the exponential. Such linear attention schemes run in $O(N)$ time,
but still suffer a noticeable performance gap between transformer models.

\noindent
\textbf{Recent Modern Linear Attention}
Recent linear attention variants dispense with softmax normalization and utilize identity feature maps $\phi(x)=x$, yielding the formulation $\mathbf{o}_{t}=\mathbf{q}_{t} \mathbf{S}_{t}$, where the KV state $\mathbf{o}_{t}=\sum_{i=1}^{t} \mathbf{k}_{i}^{T}\mathbf{v}_{i}$ is computed recursively as $\mathbf{S}_{t} = \mathbf{S}_{t-1} + \mathbf{k}_{t}^{T}\mathbf{v}_{t}$.
Instead of directly approximating softmax function, these models focus on enriching the state update process itself. Most models propose state transitions that take the form
\begin{equation}
    \mathbf{S}_{t} = g(\mathbf{S}_{t-1}, \mathbf{x}_{t}) + f( \mathbf{k}_{t}^{T}\mathbf{v}_{t}, \mathbf{x}_{t})
\end{equation}
The function $g(,)$ often implements input-dependent gating or forgetting, controlling the influence of the historical state $\mathbf{S}_{t-1}$~\cite{yang2023gated, peng2024rwkv6}.
The function $f(,)$ defines how the current key-value outer product contributes to the state update~\citep{mamba2, deltanet, azizi2025mambaextend, ye2025lamb}. 
Both functions can accept the input $\mathbf{x}_{t}$ to make them input-dependent.
These approaches leverage the efficiency of linear recurrence while introducing sophisticated mechanisms for state evolution and information flow, resulting in increasingly powerful sequence modeling capabilities.
However, pre-training demands significant computational resources hindering widespread adoption.

\noindent
\textbf{Linearization of Attention}
To reduce the training costs, recent efforts try to leverage pre-trained transformer-based LLMs to obtain computationally efficient models with linear complexity. Hedgehog~\citep{zhang2024Hedgehog} introduced a learned linear feature mapping ($\phi$) to approximate softmax attention. Expanding on this approach, LoLCATs~\citep{zhang2025lolcats} integrated sliding-window attention to improve performance further. However, these methods still struggle to maintain strong performance on tasks involving significantly longer contexts than seen during distillation.

Another approach, MambaInLlama~\citep{junxiongdaniele2024mambainllama}, applied progressive distillation along with supervised fine-tuning~\citep{kim2016sequencelevelknowledgedistillation, rofly2025} and directed preference optimization~\citep{rafailov2023direct} to distill Mamba models from Llama3 models. Despite achieving strong performance at extended lengths up to 38K tokens, it still retains 50\% of self-attention and requires substantial training data ($\sim$20B tokens).
SUPRA~\citep{mercat2024linearizing} pursued a different strategy, modifying LLaMA3 models into architectures similar to RetNet~\citep{sun2023retentive}, but still required extensive additional training ($\sim$100B tokens) to extend effective context handling to 32K tokens. Thus, SUPRA does not fully resolve the substantial overhead associated with long-context model training.

\section{Algorithmic Motivation}

A fundamental distinction between softmax attention and its linear approximations lies in the normalization mechanism and the role of the exponential function. Softmax attention weights, $\{w_{ti}\propto \exp(\mathbf{q}_{t} \mathbf{k}_{i}^{T})\}$, are normalized by a sum over all keys for a given query. This inherently creates a competitive dynamic: the $\exp$ function sharply amplifies scores for keys highly similar to the query relative to others, effectively focusing attention mass on the most relevant tokens~\citep{vaswani2023attention}.
Although self-attention is renowned for modeling global dependencies, much of its effectiveness also stems from capturing local context (e.g., adjacent words or neighboring pixels). The softmax mechanism excels here as its exponential weighting amplifies locally coherent patterns (like increasing the relevance of "New York" when queried with "York"), effectively capturing correlations abundant in natural sequences. 
Empirical evidence~\citep{han2024bridging}, including the disproportionate performance drop observed when masking local versus random tokens, further demonstrates that softmax models leverage local context more effectively than typical linear variants.

In contrast, linear attention often employs feature maps $\phi$ applied independently to queries and keys, yielding attention weights $\tilde{w}_{ti}\propto \phi (\mathbf{q}_{t})\phi(\mathbf{k}_{i})^{T}$.
The normalization term $\sum_{j}\phi(\mathbf{q}_{t})\phi(\mathbf{k}_{j})^{T}$ simply aggregates transformed key features without the strong competitive effect of the softmax denominator. 
This can lead to limitations such as attention dilution or oversmoothing, where attention distributions become less peaked \citep{qin-etal-2022-devil}. More critically, this formulation struggles to replicate the strong local modeling capabilities inherent in softmax attention \citep{han2024bridging}.
This weaker local modeling is a key drawback of naive linear attention. Meanwhile, the potentially low-rank nature of the $\phi(\mathbf{Q})\phi(\mathbf{K})^T$ kernel can make it insensitive to crucial local variations and contextual nuances. Two distinct queries $q$ and $q'$, even if originating from different local contexts, might yield similar $\phi(q)\approx \phi(q')$ and thus nearly identical attention patterns, a failure mode less likely in softmax due to its context-sensitive normalization.

\section{Proposed Method}
\subsection{Integration of Causal Conv1D Layer}

\begin{figure}
    \centering
    \includegraphics[width=0.95\linewidth]{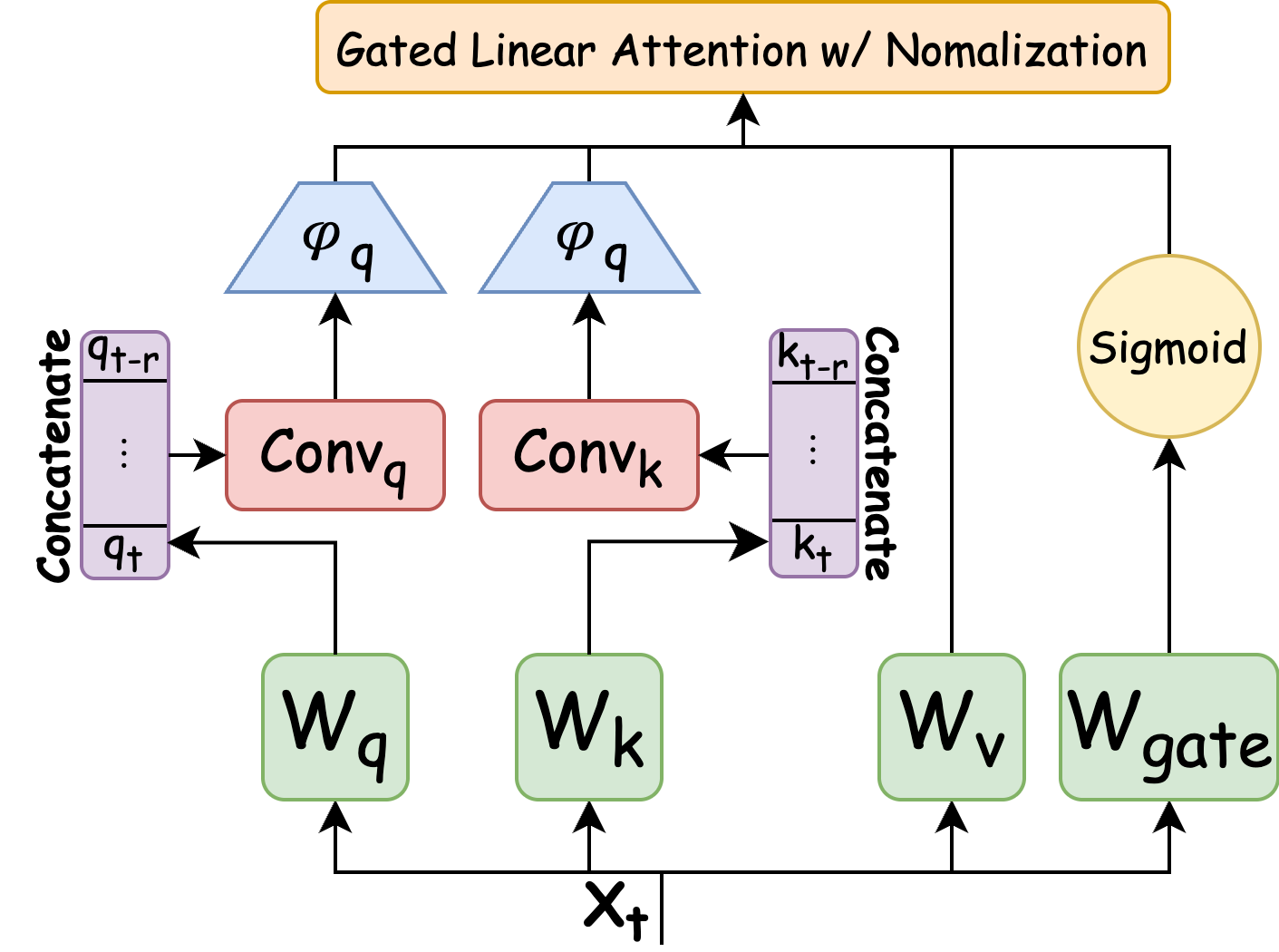}
    \caption{The overall structure of LAWCAT. We use Casual Conv1D with a kernel size of $r+1$ for visualization, and use $\mathbf{x}_t$ to represent the $t$-th token of input $\mathbf{x}$.} 
    \vspace{-4mm}
    \label{fig:framework}
  \end{figure}

To address this deficiency in capturing local structure, we propose to integrate Conv1D layers to make the function $\phi$ more context-sensitive.
In particular, as shown in Fig.~\ref{fig:framework}, our LAWCAT adds a causal depthwise convolution layer with kernel size of $r+1$ (In practice, we set kernel size as 4) for the query and key, respectively, to introduce an inductive bias for locality into the attention approximation. 
Since this convolution layer is causal, it only uses the tokens before the current token to calculate the convolution. 
Similar to the KV cache mechanism, we can cache the query and key from the previous tokens, but the size of the cache is limited by the kernel size.
After the Conv1D layer, similar to the LoLCATs~\citep{zhang2025lolcats}, we use a linear layer with nonlinearity to project the query and key but we make the query and key to share the same linear projection, which makes the approximation align with the Eq.~\ref{eq:la:1} which use the same function $\phi$ for the query and value. 

A causal Conv1D with kernel size of $r+1$ will compute each transformed query $\tilde{\mathbf{q}}_{t}= f_{\theta}\big ([\mathbf{q}_{t-r},...,\mathbf{q}_{t},] \big)$ as a function of a local window of the original queries (and similarly $\tilde{\mathbf{k}}_{i}$ from neighboring keys). 
The convolution is linear (a depthwise 1D convolution is essentially a weighted moving average of the inputs in a local window) and we do not include nonlinearities between this Conv1D layer and the subsequent linear layer. 
In the simplest case, let $\tilde{\mathbf{q}}_{t}= \sum_{\delta=0}^{r}w_{\delta}\,\mathbf{q}_{t-\delta}$ with some learnable weights $w_{0 \dots r}$ (and similarly $\tilde{\mathbf{k}}_{i}$). This operation mixes local token information into the queries and keys before they interact. Intuitively, the convolution can be seen as smoothing and enriching the representations of $Q$ and $K$ with their neighbors' features. As a result, the kernel function $\phi$ (now applied on $\tilde{\mathbf{q}}_{t}$ and $\tilde{\mathbf{k}}_{i}$) is no longer a purely pointwise function – it has implicit awareness of nearby tokens. 

The similarity between $\tilde{\mathbf{q}}_{t}$ and $\tilde{\mathbf{k}}_{i}$ in the conv-augmented space includes contributions from keys in the neighborhood of $i$ as well. In effect, a key $\mathbf{k}_{i}$ will receive a high weight $\tilde{w}_{ti}$ not only if it aligns with the query, but also if its previous neighbors $\mathbf{k}_{i-1}, \mathbf{k}_{i-2},\dots$ align with the query. This property can help the linear attention mimic what softmax does: softmax will give a group of similar, adjacent keys a collectively high weight if the query matches that group. The Conv1D introduces an implicit group-aware effect as neighboring keys support each other's relevance. Conversely, if a key $\mathbf{k}_{i}$ is an outlier among irrelevant neighbors, $\tilde{\mathbf{k}}_{i}$ may be diluted by its neighbors (lowering $\phi (\tilde{\mathbf{k}}_{i})$'s dot with $\phi (\tilde{\mathbf{q}}_{t})$), which is analogous to how an isolated token might not stand out as strongly under softmax normalization if its similarity is not much larger than others. In summary, the convolution provides a localized smoothing and sharpening of attention scores: smoothing because it aggregates local information mitigating random fluctuations, and sharpening because a coherent local pattern (several similar tokens in a window) will amplify each token's effective feature.

\textbf{The advantages of the Conv1D layer over the sliding window attention} On the other hand, LoLCATs ~\cite{zhang2025lolcats} proposed to use a sliding window attention to improve the overall performance. Specifically, for the tokens within the window, they will calculate the softmax attention score, and for the tokens outside the window, they will use the linear attention score, and use the weighted sum as the final attention score. However, this strategy does not attempt to improve the linear attention component's handling of local interactions, so retrieving information located far beyond the window relies solely on the capabilities of the linear attention component. If the linear attention part fails to accurately capture these long-range dependencies or lacks the necessary precision, the model's ability to retrieve distant information will be significantly impaired, leading to performance degradation as the distance to the target information increases.

In contrast, our LAWCAT produces one cohesive attention output that naturally balances local and global information which gets rid of the difficulties of balancing the linear and softmax attention. Additionally, by blending neighboring tokens' features, LAWCAT can reduce the risk of query confusion (non-injectivity) and increase the effective rank of the attention, leading to a better approximation of full attention~\citep{fan2024breaking}. Moreover, LAWCAT encodes a prior that nearby tokens are related, which can improve generalization on sequences with local structure helping combat the tendency of full attention to overfit noise.

\subsection{GLA with Normalization}
Another key design choice in our architecture is the adoption of Gated Linear Attention (GLA), which adds additional linear layers followed by a sigmoid activation to enable input-dependent gating.
Unlike~\citet{yang2023gated}, who omit the normalization term in Eq.~\ref{eq:la:1}, we find that retaining this normalization is critical for both training stability and final performance, particularly in the context of distillation from a standard Transformer model.

Let $\mathbf{G}_t$ denote the forget gate and $\mathbf{S}_t$ represent the key-value (KV) state of the linear attention mechanism at time step $t$. For notational simplicity, we denote the projected query and key of the $i$-th token, after Conv1D and linear projection, as $\mathbf{\dot{q}}_i$ and $\mathbf{\dot{k}}_i$, respectively.
Then, the KV state update equation in GLA can be described as:
  \begin{equation}
    \mathbf{S}_{t}= \mathbf{G}_{t}\odot \mathbf{S}_{t-1}+ \mathbf{\dot{k}}_{t}^{T}\mathbf{v}_{t}
  \end{equation}
\noindent
where $\odot$ is the Hadamard product. Letting the $\mathbf{S}_{0}=0$, we can rewrite the above equation as:
  \begin{equation}
    \mathbf{S}_{t}= \sum_{i=1}^{t}\left[\left( \bigodot_{z=i+1}^{t}\mathbf{G}_{z}\right) \odot \left (\mathbf{\dot{k}}_{i}^{T}\mathbf{v}_{i}\right) \right]
  \end{equation}
\noindent
and we use $\mathbf{J}_{i, t}$ to denote the consecutive Hadamard products when $i+1 \leq t$:
  \begin{equation}
    \mathbf{J}_{i, t}= \bigodot_{z=i+1}^{t}\mathbf{G}_{z}= \mathbf{G}_{t}\odot \mathbf{G}_{t-1}\odot \dots \odot \mathbf{G}_{i+1}\label{eq:la:J}
  \end{equation}
\noindent
otherwise, we set $\mathbf{J}_{i, t}= \mathbf{1}$. Finally, we get the expression of the KV state as Eq.~\ref{eq:la:state} and the attention output as Eq.~\ref{eq:la:output}:
  \begin{equation}
    \mathbf{S}_{t}= \sum_{i=1}^{t}\left[ \mathbf{J}_{i, t}\odot (\mathbf{\dot{k}}_{i}^{T}\mathbf{v}_{i}) \right] \label{eq:la:state}
  \end{equation}
  \begin{equation}
    \mathbf{o}_{t}= \frac{\mathbf{\dot{q}}_{t}\sum_{i=1}^{t}\left[ \mathbf{J}_{i, t}\odot (\mathbf{\dot{k}}_{i}^{T}\mathbf{v}_{i}) \right]}{\mathbf{\dot{q}}_{t}\sum_{j=1}^{t}\left[ \mathbf{J}_{j, t}\odot (\mathbf{\dot{k}}_{j}^{T}\mathbf{1}) \right]_{mean(1)}}\label{eq:la:output}
  \end{equation}
Note, since $\mathbf{G}_{i}\in \mathbb{R}^{d_k\times d_v}$, we also have $\mathbf{J}_{i,t}\in \mathbb{R}^{d_k\times d_v}$, but $\mathbf{k}_{i}\in \mathbb{R}^{1\times d_k}$, it makes it not straightforward to calculate the normalization of the GLA. To address this, we let the $\mathbf{k}_{i}^{T}$ times $\mathbf{1}_{1\times d_v}$ first, which actually repeat the $\mathbf{k}_{i}^{T}$ for $d_{v}$ times along the second dimension, then follow the normal GLA operation. Before the multiplication with $\mathbf{q}_{t}$, we take the mean value of the $\mathbf{J}_{i+1, t}\odot (\mathbf{k}_{i}^{T}\mathbf{1})$ along the second dimension to reduce the shape from $d_{k}\times d_{v}$ to $d_{k}\times 1$. In this way, we normalize the GLA with a similar definition as in Eq.~\ref{eq:la:1}.

  \begin{figure*}[t]
    \includegraphics[width=0.49\linewidth]{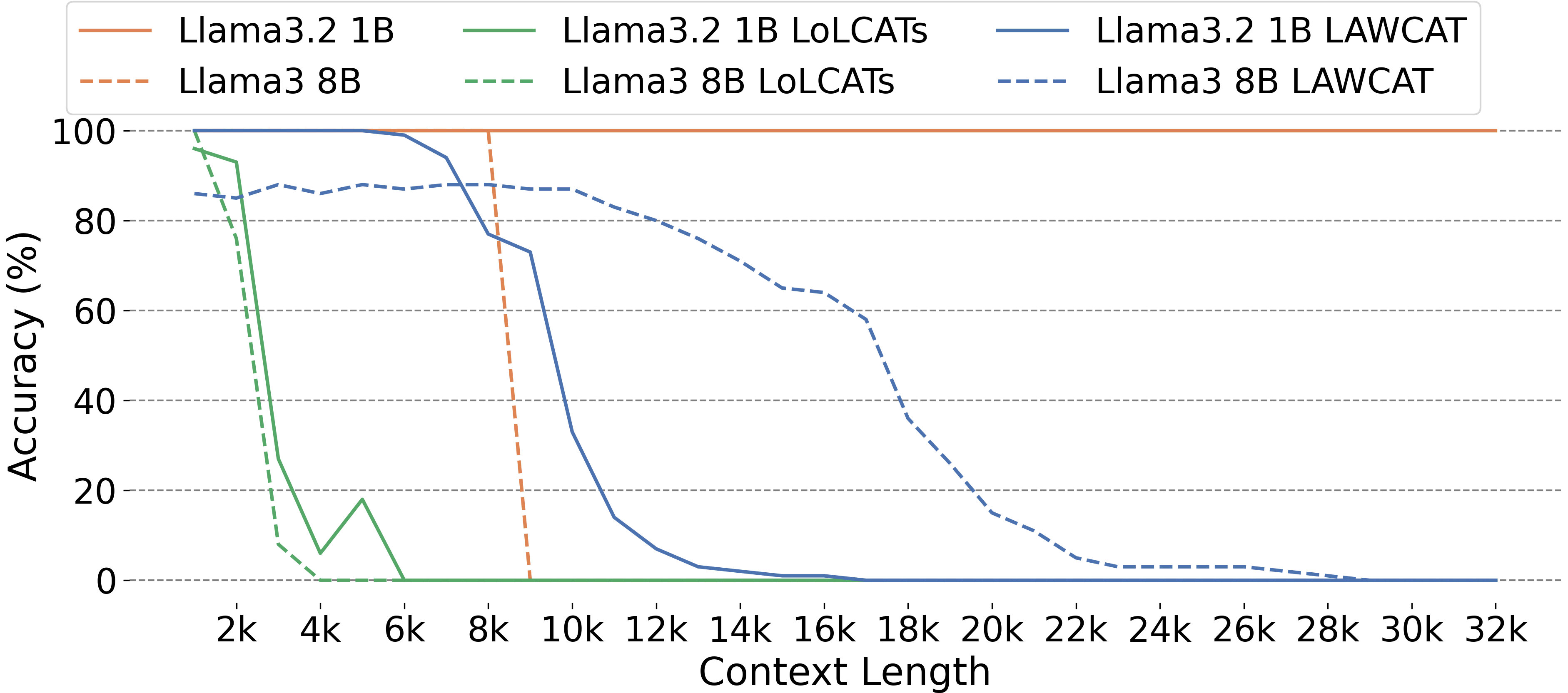}
    \hfill
    \includegraphics[width=0.49\linewidth]{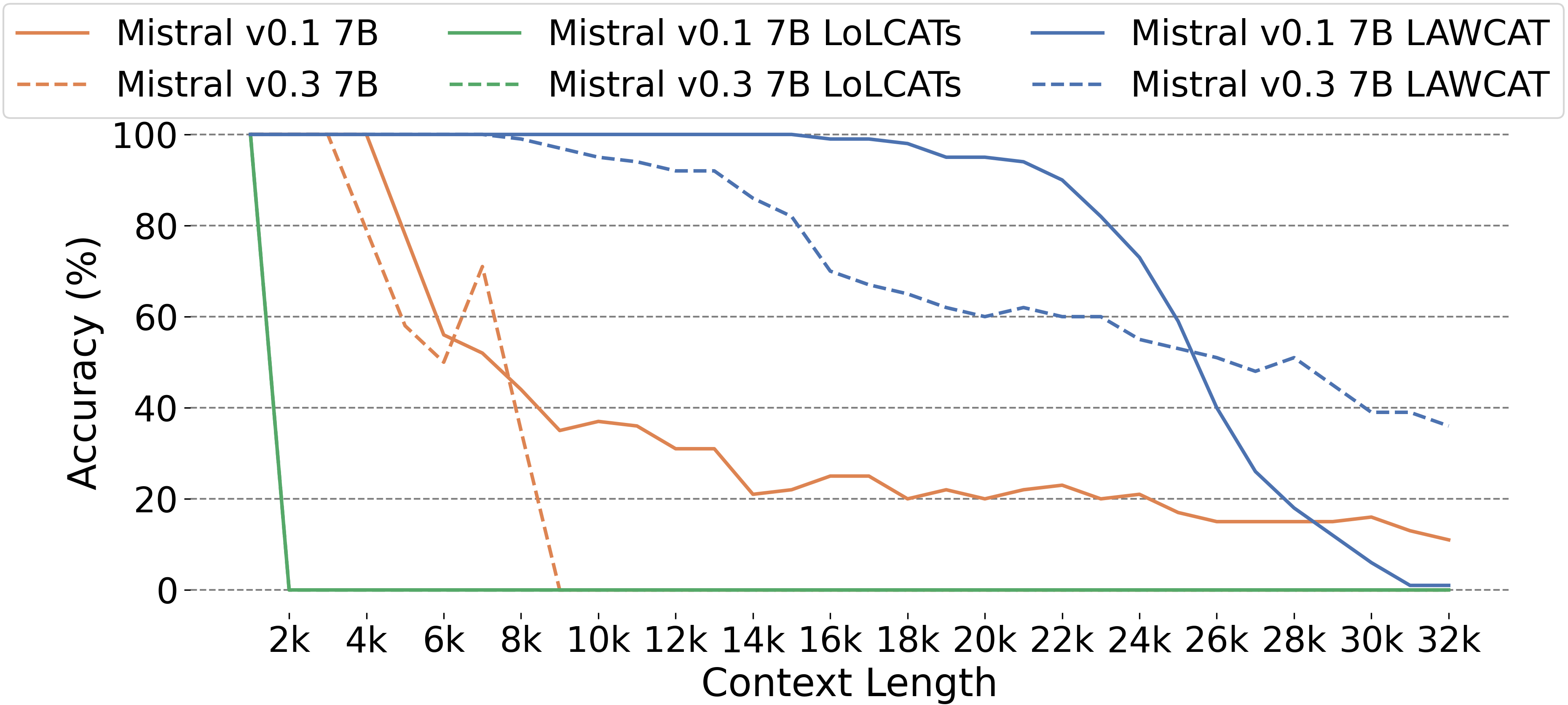}
    \caption{The comparison between the pre-trained transformer model and our converted linear attention model.}
    \label{fig:pk}
  \end{figure*}
  
\section{Experimental Results}

\subsection{Setup}

 \begin{table*}
  \small
    \centering
    \begin{tabular}{l|cccc|cccc|ccc}
      \toprule
                                          & \multicolumn{4}{c|}{S-NIAH-1}                      & \multicolumn{4}{c|}{S-NIAH-2}             & \multicolumn{3}{c}{S-NIAH-3}           \\
                                          & \multicolumn{4}{c|}{(pass-key retrieval)}          & \multicolumn{4}{c|}{(number in haystack)} & \multicolumn{3}{c}{(uuid in haystack)} \\
      \midrule
                                          & 1K                                                 & 2K                                        & 4K                                    & 8K            & 1K           & 2K            & 4K            & 8K          & 1K            & 2K          & 4K          \\
      \midrule
      Llama3.2 1B Instruct & 100 & 100 & 100 & 100 & 100 & 100 & 96 & 100 & 100 & 100 & 100 \\
      DeltaNet-1.3B                       & 97.4                                               & 96.8                                      & 99.0                                  & \textbf{98.8} & 98.4         & 45.6          & 18.6          & 14.4        & 85.2          & 47.0        & 22.4        \\
      Mamba2-1.3B                         & 99.2                                               & 98.8                                      & 65.4                                  & 30.4          & 99.4         & 98.8          & 56.2          & 17.0        & 64.4          & 47.6        & 4.6         \\
      Gated DeltaNet-1.3B                 & 98.4                                               & 88.4                                      & 91.4                                  & 91.8          & 100        & \textbf{99.8} & \textbf{92.2} & 29.6        & \textbf{86.6} & 84.2        & 27.6        \\
      \midrule
      \textbf{\textit{Distilled Model}}   & \multicolumn{11}{c}{Pre-trained Model: Llama3.2-1B-Instruct} \\
      LoLCATs                             & 100                                                & 84                                        & 0                                     & 0             & 84           & 44            & 0             & 0           & 72            & 24          & 0           \\
      Ours                                & \textbf{100}                                       & \textbf{100}                              & \textbf{100}                          & 80            & \textbf{100} & 96            & 88            & \textbf{48} & 56            & 44          & 24          \\
      \midrule
      & \multicolumn{11}{c}{Pre-trained Model: Llama3-8B-Instruct}   \\
      LoLCATs                             & 100                                                & 4                                         & 0                                     & 0             & 92           & 4             & 0             & 0           & 84            & 24          & 0           \\
      Ours                                & 100                                                & 100                                       & 96                                    & 80            & 100          & 92            & 84            & 32          & 80            & \textbf{88} & \textbf{60} \\
      \midrule
        & \multicolumn{11}{c}{Pre-trained Model: Mistral v0.1 7B} \\
      LoLCATs  & 100 & 100 & 100 & 100	& 100 & 96 & 64 & 0	& 72 & 72 & 16                         \\
      \midrule
      & \multicolumn{11}{c}{Pre-trained Model: Mistral v0.3 7B} \\
      LoLCATs  & 100 & 100 & 100 & 100	& 100 & 88 & 60 & 16	& 68 & 24 & 12                         \\
    \bottomrule
    \end{tabular}
    \caption{Performance comparison across different models on S-NIAH 1, 2, and 3 tasks. }
    \label{tab:niah_results}
  \end{table*}

Our training followed the two-stage methodology of LoLCATs~\citep{zhang2025lolcats}: an initial distillation stage employing only layer-wise MSE loss, followed by LoRA fine-tuning~\citep{hu2022lora}. In total, for Passkey Retrieval~\citep{mohtashami2023passkey}, S-NIAH~\citep{hsieh2024ruler}, and BABILong~\citep{NEURIPS2024_babilong} tasks, we use 93M, 31M, and 103M tokens, respectively, with maximum input lengths of 1162, 1223, and 1300 tokens. More details regarding training datasets and training configurations are provided in Appendix~\ref{appendix:1} and~\ref{appendix:2}.
Standard pre-trained models typically leverage over 100B tokens to attain competitive performance. Compared with them, our approach requires less than 0.1\% of such pre-training tokens.

\subsection{Results on Passkey Retrieval Task}

As shown in Fig.~\ref{fig:pk}, the pre-trained LLaMA3 8B~\citep{grattafiori2024llama3herdmodels} maintains perfect accuracy from 1K to 8K tokens, but drops to zero beyond 9K, exceeding its training context length.
In contrast, our method extends the effective context window: even when distilled and fine-tuned solely on 1K-length sequences, the resulting model retains competitive performance up to 12K tokens. Similarly, for the pre-trained LLaMA3.2 1B model, which performs reliably up to 32K tokens, our approach preserves strong retrieval accuracy up to 7K tokens. Compared to the LoLCATs method, whose effectiveness is limited to 1K for LLaMA3 8B and 1–2K for LLaMA3.2 1B, our method demonstrates substantially superior preservation of long-context retrieval capabilities.

We also evaluate our approach on the Mistral v0.1 and v0.3 7B models~\citep{jiang2023mistral7b}, which fail to achieve 100\% accuracy even within 8K, likely due to their limited optimization for retrieval tasks. Despite this, our linearized Mistral v0.3 7B model retains good accuracy up to 15K tokens, and the linearized Mistral v0.1 7B model extends this further, achieving over 90\% accuracy up to 22K tokens. 
In contrast, the LoLCATs-distilled Mistral models reach peak performance at only 1K tokens and rapidly degrade to zero beyond 2K tokens. These results  highlight the robustness and generalizability of our method in preserving long-context capabilities across various model architectures.

\subsection{Results on S-NIAH Benchmark}
Beyond the Passkey Retrieval task, we evaluate our method on the more challenging S-NIAH 1-3 task from the RULER benchmark~\citep{hsieh2024ruler}. For a comprehensive comparison, we include results from several SOTA pre-trained recurrent models, DeltaNet~\citep{deltanet}, Mamba2~\citep{mamba2}, and Gated DeltaNet~\citep{gateddelta}. All pre-trained models are trained on 100B tokens, and the training length is 4K. And the results are sourced from ~\citet{gateddelta}.

As shown in Table~\ref{tab:niah_results}, on the S-NIAH 1 task, our model achieves 100\% accuracy from 1K to 4K tokens, surpassing all listed SOTA recurrent models. Even at 8K, it remains competitive, despite being trained on sequences more than 3$\times$ shorter  (1223 vs. 4000 tokens) and using only 0.03\% of the data (31M vs. 100B tokens).
On the S-NIAH 2 task, LAWCAT achieves comparable performance up to 4K and maintains a strong 48\% accuracy at 8K—substantially outperforming other models and demonstrating robust generalization to longer contexts. 
While performance on the more complex S-NIAH 3 task lags slightly behind others, we hypothesize this is due to the limited diversity in our training data. 
LoLCATs only maintain good results around 1K for all S-NIAH tasks, which is consistent with the results on the Passkey Retrieval task.

The results related to the Llama3 8B and Mistral series models demonstrate similar trends; our model demonstrates superior robustness to increasing input lengths, with notably smaller performance drop compared to LoLCATs and other SOTA pre-trained recurrent models.

\begin{figure}[!htbp]
    \centering
    \includegraphics[width=0.95\linewidth]{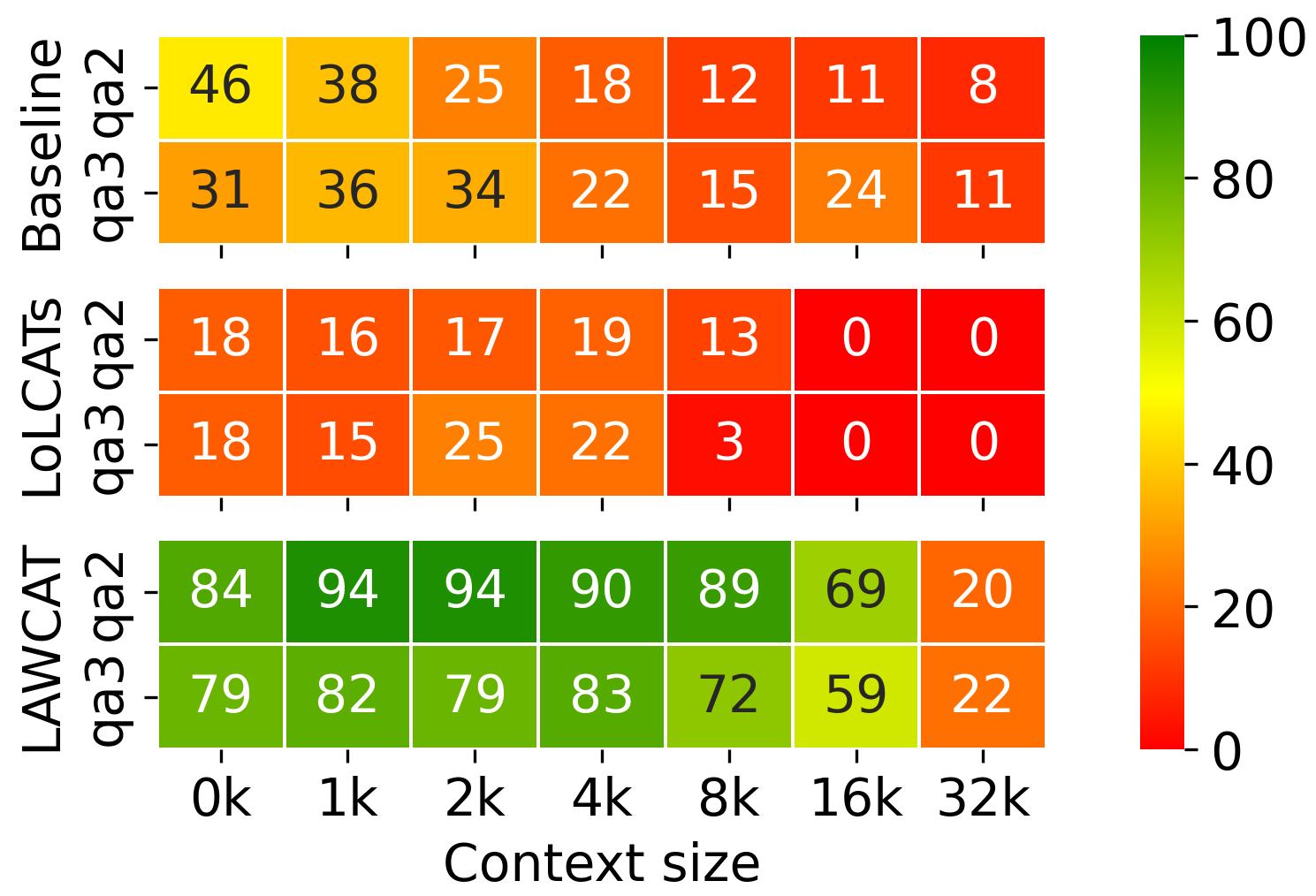}
    \caption{Comparison of accuracy on QA2\&3 from BABILong benchmark between Llama 3.2 1B (top), LoLCATs (middle), and LAWCAT (bottom).}
    \label{fig:babilong}
    \vspace{-3mm}
\end{figure}
\subsection{Results on BABILong benchmark}
Besides retrieval capabilities, we assessed LAWCAT's performance on complex reasoning tasks using the BABILong benchmark \citep{NEURIPS2024_babilong}, specifically focusing on multi-hop reasoning tasks, QA2 (two supporting facts) and QA3 (three supporting facts). The results, presented in Figure~\ref{fig:babilong}, demonstrate a significant advantage for our distilled and fine-tuned LAWCAT variant compared with the original Llama3.2 1B model. Across the evaluated context lengths (0K to 32K tokens), LAWCAT consistently maintains accuracy levels more than double those of the baseline Llama3.2 1B model for both QA2 and QA3 tasks.

This notable enhancement in multi-hop reasoning performance underscores the robustness of the LAWCAT framework, particularly highlighting its capability to overcome inherent limitations observed in the foundational pretrained model. 
Moreover, compared with LoLCATs, which failed in this task, the results demonstrate LAWCAT’s capacity not merely for transferring existing knowledge but also for significantly enhancing complex reasoning skills within extensive contexts. 
We also evaluate the Llama3 8B and Mistral series models on the BABILong benchmark, more details are shown in Appendix~\ref{appendix:babilong}.

\subsection{Efficiency Comparison}
To evaluate computational performance, we benchmarked the pre-fill latency of our Llama3.2 1B LAWCAT model against a Llama3.2 1B model incorporating FlashAttention-2 (FA2) \citep{dao2023flashattention2} and the LoLCATs variant, using one NVIDIA RTX A6000 GPU. For a comprehensive comparison, we also include results from SOTA recurrent models, specifically GLA 1.3B and Mamba 1 1.3B.

\begin{figure}[!htbp]
    \centering
    \includegraphics[width=0.9\linewidth]{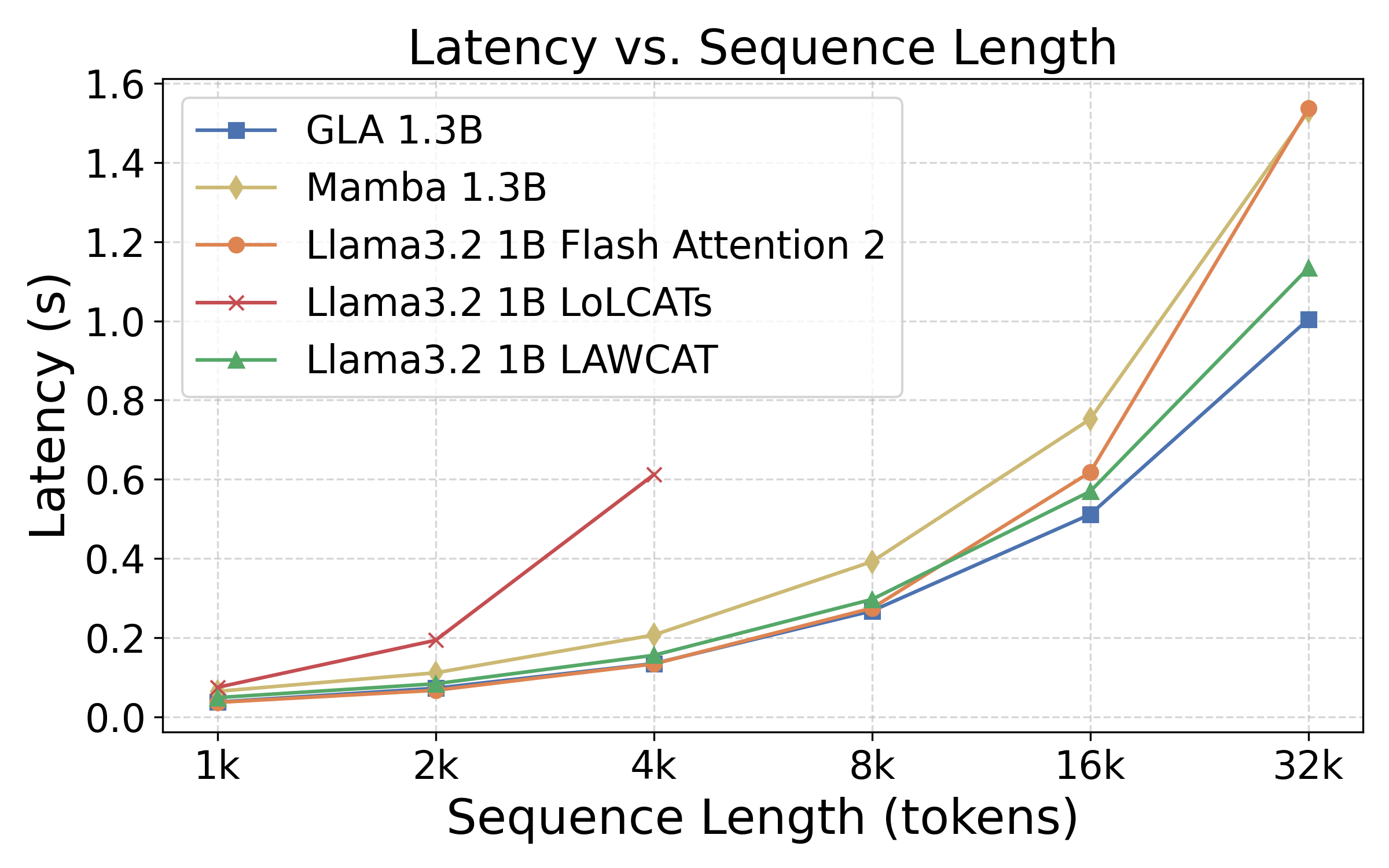}
    \caption{Comparison of prefill-stage latency among 5 different models. Note, LoLCATs runs out of memory for sequence lengths exceeding 8K tokens.}
    \label{fig:latency}
\end{figure}

As illustrated in Fig.~\ref{fig:latency}, LAWCAT incurs slightly higher prefill latency than LLaMA3.2 1B with FA2 for sequences shorter than 8K tokens. Beyond this point, however, LAWCAT achieves significantly lower latency, with the gap widening as sequence length increases.
In contrast, the naive implementation of LoLCATs (i.e., LoLCATs without the ThunderKittens kernel~\citep{thunderkittens2024}) exhibits substantially higher latency and greater GPU memory consumption. 

When compared to the GLA 1.3B model, LAWCAT's inclusion of a Conv1D and linear layer, and GLA normalization contributes to a slightly higher latency. 
Notably, LAWCAT consistently outperforms the Mamba 1 1.3B model, which also incorporates a Conv1D layer, across all evaluated sequence lengths.

\subsection{Ablation Study and Discussion}
We conducted a comprehensive ablation study to systematically evaluate the contribution of each component in the proposed LAWCAT framework and discuss the effect of prevalent techniques such as Rotary Position Embeddings (RoPE) and Stochastic Weight Averaging (SWA). 
Unless otherwise specified, all experiments are based on models distilled and fine-tuned from the pre-trained LLaMA3.2-1B using training data with a sequence length of $\sim$1K tokens. 
The ablation analysis is structured around the following key questions:\\

\begin{table}[!htbp]
    \small
    \centering
    \begin{tabular}{lccccccc}
    \toprule
         & 1K & 2K $\sim$ 6K & 7K  & 8K  & 9K  & 10K & 11K \\
    \midrule
    NN   & 0.9& 0            & 0 & 0 & 0 & 0 & 0 \\
    NC   & 1.0& 1.0          & 1.0 & 0.8 & 0.9 & 0.9 & 0.4 \\
    O    & 1.0& 1.0          & 0.9 & 0.9 & 0.8 & 0.5 & 0.2 \\
    \bottomrule
    \end{tabular}
    \caption{Accuracy on the passkey retrieval task from 1K to 11K. "NN" means removing the GLA normalization from LAWCAT, "NC" means removing the Conv1D layer, and "O" means the complete LAWCAT model.}
    \label{tab:ablation:pk}
\end{table}

\noindent
\textbf{Do we need the GLA normalization and Conv1D across tokens?}
To assess the contribution of key components within our proposed LAWCAT framework, we conduct an ablation study focusing on the GLA normalization and Conv1D layer.

As presented in Table~\ref{tab:ablation:pk}, removing GLA normalization significantly degrades performance on passkey retrieval tasks involving longer contexts, although the model maintains reasonable accuracy at the 1K-length scale. This finding supports our hypothesis that normalization is crucial for generalization. By constraining the GLA attention output to remain a normalized weighted sum of the values ($\mathbf{v}_i$) during layer-wise distillation (via MSE loss minimization against self-attention outputs), the normalization term mitigates overfitting to shorter sequence lengths and enhances the model's ability to preserve performance as context length increases.

On the other hand, the results in Table~\ref{tab:ablation:pk} suggest that the Conv1D layer offers limited benefit for the relatively simpler passkey retrieval task. We attribute this to the nature of the task, where much of the context contains redundant or useless information; applying Conv1D over such sequences does not necessarily facilitate the extraction of more salient features for identifying and memorizing the passkey. However, its importance becomes evident in more complex scenarios. Table~\ref{tab:ablation:niah} reveals a substantial performance gap on the NIAH task between the full LAWCAT model and its variant without the Conv1D layer. Notably, without Conv1D, LAWCAT fails to retrieve the correct UUID in all tested cases. This stark difference underscores the critical role of the Conv1D layer in capturing dependencies and memorizing information across tokens, particularly when dealing with longer sequences and more intricate information retrieval demands.

\begin{table}[!htbp]
  \small
  \setlength{\tabcolsep}{3pt}
    \centering
    \begin{tabular}{l|cccc|cccc|ccc}
      \toprule
                                          & \multicolumn{4}{c|}{S-NIAH-1}                      & \multicolumn{4}{c|}{S-NIAH-2}             & \multicolumn{3}{c}{S-NIAH-3}           \\
      \midrule
                                          & 1K                                                 & 2K                                        & 4K                                    & 8K            & 1K           & 2K            & 4K            & 8K          & 1K            & 2K          & 4K          \\
      \midrule
      NC                             & 96  & 96 & 96 & 8  & 80   & 80     & 80     & 36     & 0   & 0  & 0  \\
      O                                & 100                                       & 100                              & 100                          & 80            & 100 & 96            & 88            & 48 & 56            & 44          & 24          \\
    \bottomrule
    \end{tabular}
    \caption{Accuracy on S-NIAH 1\&2\&3 tasks. "NC" means removing the Conv1D layer from LAWCAT, and "O" means the complete LAWCAT model. }
    \label{tab:ablation:niah}
  \end{table}

\noindent
\textbf{Do we need the RoPE and SWA?} 
The necessity of explicit positional encodings like RoPE in linear attention is an open question, with varied adoption across models. Our ablation study (Table \ref{tab:ablation:rope} in Appendix~\ref{appendix:ab}) on LoLCATs and LAWCAT models addresses this. For LoLCATs, removing RoPE decreased performance on shorter contexts (1K-3K tokens) but slightly improved it on longer ones (4K-8K). This effect was more stark for our LAWCAT: the variant without RoPE maintained high accuracy up to 8K tokens, whereas the RoPE-equipped variant's performance collapsed beyond 3K tokens. These findings suggest that RoPE, while potentially aiding short-sequence modeling, introduces biases detrimental to long-context generalization, supporting the idea that recurrent formulations can inherently capture sequence order.

We also investigated the role of SWA (Table \ref{tab:ablation:swa} in Appendix~\ref{appendix:ab}). On long-context passkey retrieval, removing SWA significantly impaired LoLCATs, confirming its reliance on this mechanism. Conversely, LAWCAT performed better without SWA, especially at longer contexts; we hypothesize this is due to the difficulty in balancing contributions from the GLA and SWA, particularly when generalizing from shorter training to longer evaluation sequences. Additional results and analysis on the impact of SWA on standard short-context benchmarks (LM Eval, Table \ref{tab:ablation:swa}) are presented in Appendix~\ref{appendix:ab}.


\section{Conclusions}
We introduced LAWCAT (Linear Attention with Convolution Across Time), a novel and efficient distillation framework that strategically integrates a causal Conv1D layer to bolster local dependency modeling and employs a normalized gated linear attention mechanism for robust generalization across varying sequence lengths. 
Our comprehensive experiments validate the effectiveness of LAWCAT. Distilling models like Mistral v0.1 7B and Llama3.2 1B yielded linear attention variants capable of remarkable long-context performance, achieving over 90\% accuracy on passkey retrieval tasks up to 22K tokens and matching SOTA recurrent models on NIAH benchmarks, all while using minimal training data and shorter sequence lengths during distillation. 
Furthermore, the resulting LAWCAT models demonstrate practical efficiency gains, exhibiting faster prefill latency than highly optimized FlashAttention-2 implementations for sequences exceeding 8K tokens.

LAWCAT offers a practical and resource-efficient pathway for developing and deploying long-context models by leveraging existing pretrained transformers. This approach helps democratize access to moderately long-sequence modeling capabilities, particularly beneficial for resource-constrained edge environments or rapid prototyping. Furthermore, due to its causal and linear nature, LAWCAT is particularly attractive for streaming inference, opening new opportunities for low-latency, on-device reasoning. 

\clearpage

\section*{Limitations}
While LAWCAT demonstrates strong long-context performance,
on some complex benchmarks like S-NIAH-3 (1K) as shown in Table~\ref{tab:niah_results} and  LM Evaluation Harness tasks~\citep{eval-harness} as shown in Table~\ref{tab:ablation:swa}, purely distilled linear attention models may not fully match the performance of similarly sized transformer models or the linear attention model with sliding window attention (SWA). Although incorporating the SWA can bridge this gap on shorter sequences, as observed in our experiments, it potentially compromises the model's effectiveness on tasks requiring very long context generalization. 
This presents an interesting avenue for future research: developing dynamic mechanisms that can adaptively balance the global reach of linear attention with the local strength often captured by windowed approaches, perhaps conditioned on input characteristics or sequence length. Alternatively, exploring the integration of more sophisticated linear attention designs like Mixture-of-Memories~\citep{du2025mom} might further mitigate the performance gap on standard benchmarks, although this could introduce new optimization challenges. Addressing this trade-off between benchmark performance and long-context fidelity remains a key direction for advancing efficient attention mechanisms.


\section*{Acknowledgments}
This work was supported in part by Amazon through the 2024-2025 Amazon ML fellowship. 

  \bibliography{custom}
\clearpage
  \appendix

  \section{Appendix}
  \subsection{Training Dataset}
  \label{appendix:1}
  All datasets are in English.
  \textbf{Passkey Retrieval}
  We use the same format as ~\cite{mohtashami2023passkey} to generate 10K samples to build the training dataset and 1K samples for validation. For the tokenizer of the Llama3.2 model, the length of each sample is 1162 tokens, and for the tokenizer of the Mistral model, the length is 1171 tokens. \\
  \textbf{S-NIAH 1\&2\&3}
  We use the same format as ~\cite{hsieh2024ruler} to generate 8456 samples to build the training dataset and 1K samples for validation. We used the BookSum dataset \citep{kryscinski2021booksum} as the contextual 'haystack', different from the Paul Graham Essay Collection Dataset~\cite{goel2023paulgraham} in evaluation.  
  For the tokenizer of the Llama3.2 model, the maximum length is 1231, the minimum length is 675, the average length is 785.50, the mode length is 768, and the std is 49. We also found that adding additional 1K data with the S-NIAH-1 format can improve the overall performance. For these 1K data, the maximum length is 1223, the minimum length is 1222, the average length is 1222.92, the mode length is 1223, and the std is 0.27.\\
  \textbf{BABILong QA2\&QA3} For the BABILong benchmark~\citep{NEURIPS2024_babilong}, training and validation data were generated on-the-fly per the original setup. Specifically, there are 10K samples for the training data of QA2 and 9868 samples for the training data of QA3. 
  Different from the curriculum learning strategy in the original paper, we simplified the training by fine-tuning solely on sequences of length 1300 with a mixed format of QA2 and QA3.

  \subsection{Training Configuration}
  \label{appendix:2}

All experiments were implemented using the Flash Linear Attention framework~\citep{yang2024fla}. 
For the Passkey Retrieval task, we distilled and fine-tuned the LLaMA3.2 1B model over 4 epochs each, totaling $1162\times10,000\times4\times2=92.96M$ training tokens. 
For larger models (LLaMA3 8B, Mistral v0.1 and v0.3 7B), distillation and fine-tuning were performed for 2 epochs each, resulting in $1162\times10,000\times2\times2=92.96M$ tokens for LLaMA and $1171\times10,000\times2\times2=93.68M$ tokens for Mistral. 
Similarly, for the NIAH tasks, we distilled and fine-tuned for 2 epochs, yielding approximately $(785.5\times8456 + 1222.9\times1000)\times2\times2\approx31.46M$ tokens. 
For the BABILong tasks (QA2 and QA3), datasets were generated dynamically; thus, exact token counts varied. With a maximum sequence length of 1300 tokens, the total training tokens were under $1300\times(10,000+9,868)\times2\times2\approx103.31M$.

For the Passkey Retrieval task with Llama3.2 1B model, we use the cosine learning rate schedule, and for other models and other tasks, we all use the ReduceLROnPlateau, which reduces the learning rate when the validation loss has stopped reducing. 
For all experiments, we use the Adamw optimizer~\citep{loshchilov2018decoupled} with an initial learning rate of 0.1 for distillation, 0.0001 for fine-tuning. For the LoRA fine-tuning, the rank is 8 and the alpha is 16, and we only add LoRA module for the linear layer for the query, key, value and output.

We keep the similar training hyperparameters as \citet{zhang2025lolcats}.
Specifically, for the Passkey Retrieval task with the LLaMA3.2-1B model, we adopt a cosine learning rate schedule, while for all other models and tasks, we employ the \texttt{ReduceLROnPlateau} scheduler, which decreases the learning rate when the validation loss plateaus. All experiments use the AdamW optimizer~\citep{loshchilov2018decoupled}, with an initial learning rate of 0.1 during distillation and 1e\text{-}4 for fine-tuning. During LoRA fine-tuning, we set the rank to 8 and the scaling factor ($\alpha$) to 16, applying LoRA modules exclusively to the linear projections of the query, key, value, and output layers in the attention block.  For all experiments, we run 3 times using different random seeds and report the results from the run achieving the best overall performance.


  \subsection{Other Ablation Study}
    \label{appendix:ab}
    
\noindent 
\textbf{Do we need to share the linear and Conv layers?}
In Table~\ref{tab:ablation:share}, we analyze four distinct configurations regarding parameter sharing between linear and convolutional layers.
Our empirical results demonstrate that optimal performance is achieved when allowing $Q$ and $K$ to independently process contextual information through separate convolutional layers while sharing the subsequent linear projection. 
This architecture aligns with the theoretical foundation in Equation~\ref{eq:la:1}, where both query and key utilize the same transformation function $\phi(\cdot)$. 
The separation of convolutional processing is intuitively justified, as queries and keys serve distinct roles in attention mechanisms and therefore benefit from specialized feature extraction pathways when incorporating previous token information. However, the shared linear projection enforces consistency in the embedding space where similarity is computed, maintaining the mathematical elegance of the linear attention formulation.\\

\begin{table}[!htbp]
    \small
    \setlength{\tabcolsep}{4.5pt}
    \centering
    \begin{tabular}{llccccccccccc}
    \toprule
    C & L  & 1K & 2K & 3K & 4K & 5K & 6K & 7K & 8K & 9K \\
    \midrule
    \checkmark & \checkmark  & 1.0& 1.0& 1.0& 0.9& 0.8& 0.5& 0.1&0.1 & 0  \\ 
    \checkmark & $\times$  & 0.9& 1.0& 0.3& 0  & 0.2& 0  & 0  &0   & 0  \\ 
    $\times$ & $\times$  & 1.0& 1.0& 1.0& 0.8& 0.8& 0.6& 0.3&0.1 & 0  \\
    $\times$ & \checkmark  & 1.0& 1.0& 1.0& 1.0& 1.0& 1.0& 0.9 & 0.9 & 0.8 \\
    \bottomrule
    \end{tabular}
    \caption{Accuracy on the passkey retrieval task from 1K to 9K. "C" and "L" mean if sharing the Conv1D or linear projection between query and key, "\checkmark" means yes, and "$\times$" means no.}
    \label{tab:ablation:share}
\end{table}

\noindent
\textbf{Do we need to add RoPE to linear attention?}
Modern linear attention models exhibit diverse strategies for handling positional information. Some architectures, like GLA \citep{yang2023gated} and Gated DeltaNet \citep{gateddelta}, often omit explicit position embeddings, relying on their recurrent formulation to implicitly capture sequence order. Conversely, others, including TransNormerLLM \citep{qin2023transnormerllm} and Retentive Networks \citep{sun2023retentive}, explicitly integrate mechanisms like rotary position embeddings (RoPE) \citep{su2024roformer} or its variants. This divergence raises a question regarding knowledge distillation: when transferring knowledge from standard Transformers (which typically use position embeddings) to linear attention models, is it beneficial to retain these explicit positional signals? 
Ablation study presented in Table~\ref{tab:ablation:rope} addresses this for the  LoLCATs and LAWCAT. The results consistently indicate that incorporating RoPE actually impairs generalization performance on long-context tasks.

For LoLCATs, removing RoPE decreases performance on shorter contexts (1K-3K tokens) but slightly improves performance on longer contexts (4K-8K tokens). This effect is more pronounced in our LAWCAT architecture, where the variant without RoPE maintains high accuracy form 1K to 8K context lengths, while the RoPE variant's performance drops to zero beyond 3K tokens.

These results indicate that position embeddings constitute a critical factor limiting generalization to extended context lengths. The fixed positional encoding scheme optimized for training context appears to introduce inappropriate biases when applied to longer sequences. This insight aligns with recent research suggesting that recurrent computation inherently captures sequential dependencies without requiring explicit positional information ~\citep{gu2023mamba}.\\

\begin{table}[]
    \small
    \setlength{\tabcolsep}{4.5pt}
    \centering
    \begin{tabular}{llcccccccccc}
    \toprule
    M & PE  & 1K & 2K & 3K & 4K & 5K & 6K & 7K & 8K \\
    \midrule
    LoL- & \checkmark  & 0.9& 0.8& 0.3& 0& 0& 0& 0&0  \\ 
    CATs & $\times$  & 0.4& 0.2& 0.1& 0.2  & 0.2& 0.1  & 0.2  &0.2    \\ 
    \midrule
    LAW-& \checkmark & 1& 0.8& 0& 0  & 0& 0  & 0  &0 \\
    CAT& $\times$  & 1.0& 1.0& 1.0& 1.0& 1.0& 1.0& 0.9 & 0.9 \\
    \bottomrule
    \end{tabular}
    \caption{The test accuracy on the passkey retrieval task from 1K to 8K. "M" means model name, and "PE" means if using the RoPE for linear attention. "\checkmark" means yes, and "$\times$" means no.}
    \label{tab:ablation:rope}
\end{table}

\noindent
\textbf{Do we need sliding window attention?} 
Prior work posited SWA as a key component for enhancing linear attention model performance~\citep{zhang2025lolcats}. To rigorously evaluate this claim, we conducted ablation studies on both the original LoLCATs model and our proposed LAWCAT architecture, examining the impact of SWA across different task types and context lengths.

Our investigation first focused on long-context retrieval tasks. As shown in Table~\ref{tab:ablation:rope}, row 3, removing SWA significantly degrades the performance of the LoLCATs model, suggesting its reliance on SWA for effective passkey retrieval. Conversely, incorporating SWA into our LAWCAT model adversely affected performance, particularly as context length increased. We hypothesize that this stems from the difficulty in optimally balancing the contributions of GLA and SWA. Furthermore, achieving a balance that generalizes from shorter training sequences to longer evaluation sequences appears challenging. This suggests it is difficult to balance the weights between linear attention and softmax attention, and even if the model can make their combination fit well on training data, it may be hard to generalize it to longer context lengths. For example, the ideal weights of LA and SWA for 1K-length contexts may not be good for 8K-length contexts.

\begin{table}[]
    \small
    \setlength{\tabcolsep}{4.5pt}
    \centering
    \begin{tabular}{llcccccccccc}
    \toprule
    M & SWA  & 1K & 2K & 3K & 4K & 5K & 6K & 7K & 8K \\
    \midrule
    LoL- & \checkmark  & 0.9& 0.8& 0.3& 0& 0& 0& 0&0  \\ 
    CATs & $\times$  & 0.1 & 0.0 & 0  & 0  & 0   & 0  & 0  & 0  \\
    \midrule
    LAW- &  \checkmark & 0.5 & 0.5 & 0.2 & 0.1 & 0 & 0 & 0 & 0 \\
    CAT &  $\times$  & 1.0& 1.0& 1.0& 1.0& 1.0& 1.0& 0.9 & 0.9 \\
    \bottomrule
    \end{tabular}
    \caption{The test accuracy on the passkey retrieval task from 1K to 8K. "M" means model name, and "SWA" means if using the sliding window attention. "\checkmark" means yes, and "$\times$" means no.}
    \label{tab:ablation:swa}
\end{table}

  \begin{table}[ht]
    \centering
    \small
    \setlength{\tabcolsep}{4pt}
    \renewcommand{\arraystretch}{1.2}
    \begin{tabular}{lccccc|c}
    \toprule
      \textbf{Model}                                & \textbf{PI}      & \textbf{AE} & \textbf{AC} & \textbf{HS} & \textbf{WG} & \textbf{Avg.} \\
      \midrule
      \multicolumn{7}{c}{\textit{Pre-trained Transformer Model}} \\
      Llama 3.2 1B                                  & 74.4             & 65.5        & 35.8        & 63.7        & 60.5        & 60.0          \\
      Llama 3.2 1B$^{*}$ & 73.2 & 64.5 & 35.6 & 40.3 & 59.9 & 54.7 \\
      \midrule
      \multicolumn{7}{c}{\textit{Pre-trained Recurrent Model}} \\
      DeltaNet 1.3B                                 & 71.2             & 57.2        & 28.3        & 50.2        & 53.6        & 52.1          \\
      GLA 1.3B                                      & 71.8             & 57.2        & 26.6        & 49.8        & 53.9        & 51.9          \\
      Mamba 2 1.3B                                  & 73.2             & 64.3        & 33.3        & 59.9        & 60.9        & 58.3          \\
      \midrule
      \multicolumn{7}{c}{\textit{Distilled Model}}   \\
      LoLCATs        & 59.4 & 41.2 & 23.8 & 31.4 & 50.2 & 41.2 \\  
      LoLCATs$^{*}$  & 74.6 & 63.0 & 35.1 & \textbf{63.7} & 61.5 & 59.6 \\                
      Ours           & 71.9 & 65.0 & \textbf{36.3} & 53.1 & 52.8 & 55.8 \\                
      Ours$^{*}$     & \textbf{75.3} & \textbf{65.2} & 35.1 & 62.0 & \textbf{62.0} & \textbf{59.9} \\      
      \bottomrule
    \end{tabular}
    \caption{Comparison of various models on 5 tasks from LM EVAL benchmark. Model name with $^{*}$ indicates the model add sliding window attention with size of 64. Specially, Llama 3.2 1B $^{*}$ means the Llama 3.2 1B model only uses SWA.}
    \label{tab:ablation:alpaca}
  \end{table}


We further assessed the effect of SWA on standard short-context benchmarks using  LM Evaluation Harness tasks (LM Eval)~\citep{eval-harness} (Table~\ref{tab:ablation:alpaca}). When SWA is removed, LAWCAT substantially outperforms LoLCATs, consistent with our findings on long-context tasks. However, adding SWA (window size 64) significantly boosts performance for both models, bringing them to comparable levels (59.6 vs. 59.9 average score). 
For context, we include results from relevant baselines: the standard Llama 3.2 1B transformer and several SOTA recurrent models. Notably, a modified Llama 3.2 1B using only SWA (denoted Llama 3.2 1B$^*$) achieves competitive performance (54.7 avg.) compared to the original (60.0 avg.), albeit with a marked drop on HellaSwag. This suggests that SAW alone possesses considerable capability. The strong results of LoLCATs may therefore be primarily attributed to the inherent effectiveness of SWA, with the linear component potentially offering further gains on specific tasks like HellaSwag.

In conclusion, our ablation studies reveal a nuanced role for SWA. While its integration can substantially enhance performance on short-context tasks, potentially contingent on successful optimization of the hybrid attention mechanism, it appears detrimental to generalization capability in long-context scenarios. \\
\noindent 
\textbf{Ablation study on the rank of the gate function in GLA}
GLA models ~\citep{yang2023gated} utilize 2 consecutive low-rank linear layers to implement the gate function for efficiency, with the original models adopting a rank of 16. We investigate the impact of this rank hyperparameter in our LAWCAT models, distilled from Llama3.2 1B Instruct~\citep{grattafiori2024llama3herdmodels}, on complex S-NIAH retrieval tasks: S-NIAH-1 (pass-key), S-NIAH-2 (number), and the particularly demanding S-NIAH-3 (UUID retrieval).

As shown in Table~\ref{tab:ab:rank}, LAWCAT models with rank 16 performed well on S-NIAH-1 and S-NIAH-2 but struggled significantly on S-NIAH-3 (e.g., 12\% at 1K, 0\% at 2K). This indicates that a rank of 16 limits the forget gate's capacity for recalling long, complex items like UUIDs.
Increasing the rank to 32 substantially improved S-NIAH-3 performance (e.g., to 56\% at 1K and 44\% at 2K) while maintaining strong performance on S-NIAH-1 and S-NIAH-2. This suggests rank 32 offers a better balance of capacity for complex retrieval.

Further rank increases to 64 yielded mixed results; while improving some S-NIAH-1/2 scores at longer contexts (e.g., S-NIAH-1 8K at 92\%; S-NIAH-2 8K at 64\%), it did not consistently improve S-NIAH-3 over rank 32 (36\% vs 56\% at 1K). Ranks 128 and Full Rank (FR) generally led to performance degradation across tasks, particularly at longer contexts (e.g., rank 128 scored 36\% on S-NIAH-1 8K). We hypothesize this non-monotonic trend stems from higher ranks introducing optimization challenges due to increased parameters and a greater risk of overfitting, which can impair the forget gate's ability to learn generalizable retention patterns.

In conclusion, these findings indicate an optimal gate function rank—around 32 for our 1B parameter LAWCAT models—that balances model capacity for complex retrieval against optimization difficulties and overfitting. 

  \begin{table*}
  \small
    \centering
    \begin{tabular}{l|cccc|cccc|ccc}
      \toprule
    & \multicolumn{4}{c|}{S-NIAH-1}                      & \multicolumn{4}{c|}{S-NIAH-2}             & \multicolumn{3}{c}{S-NIAH-3}           \\
    & \multicolumn{4}{c|}{(pass-key retrieval)}          & \multicolumn{4}{c|}{(number in haystack)} & \multicolumn{3}{c}{(uuid in haystack)} \\
      \midrule
     Rank   & 1K       & 2K          & 4K        & 8K            & 1K           & 2K            & 4K            & 8K          & 1K            & 2K          & 4K          \\
      \midrule
      16 & 100 & 100 & 96 & 80 & 100 & \textbf{96} & 92 & 52 & 12 & 0 & 4 \\
      32 & 100 & 100 & 100 & 80 & 100 & \textbf{96} & 88 & 48 & \textbf{56} & \textbf{44} & \textbf{24} \\
      64 & 100 & 92 & 100 & \textbf{92} & 100 & \textbf{96} & \textbf{96} & \textbf{64} & 36 & 32 & 20 \\
      128 & 100 & 92 & 88 & 36 & 92 & 72 & 72 & 4 & 20 & 24 & 12 \\
      FR & 96 & 96 & 80 & 12 & 96 & 72 & 36 & 4 & 24 & 28 & 12 \\
    \bottomrule
    \end{tabular}
    \caption{Performance comparison across different models on S-NIAH 1, 2, and 3 tasks. }
    \label{tab:ab:rank}
  \end{table*}

  \subsection{The visualization of Attention Scores}
  To qualitatively assess the attention mechanisms, we visualize attention score heatmaps from a representative head on an example from the S-NIAH-3 task.

\begin{figure}[!htbp]
    \centering
    \includegraphics[width=0.95\linewidth] {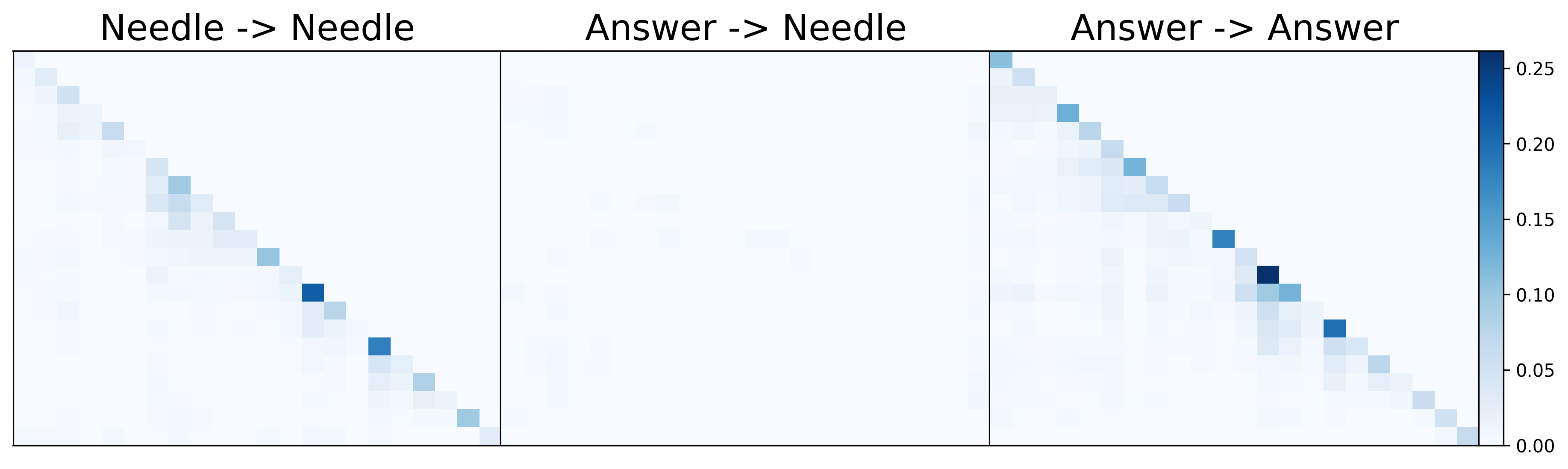} \\
    \includegraphics[width=0.95\linewidth] {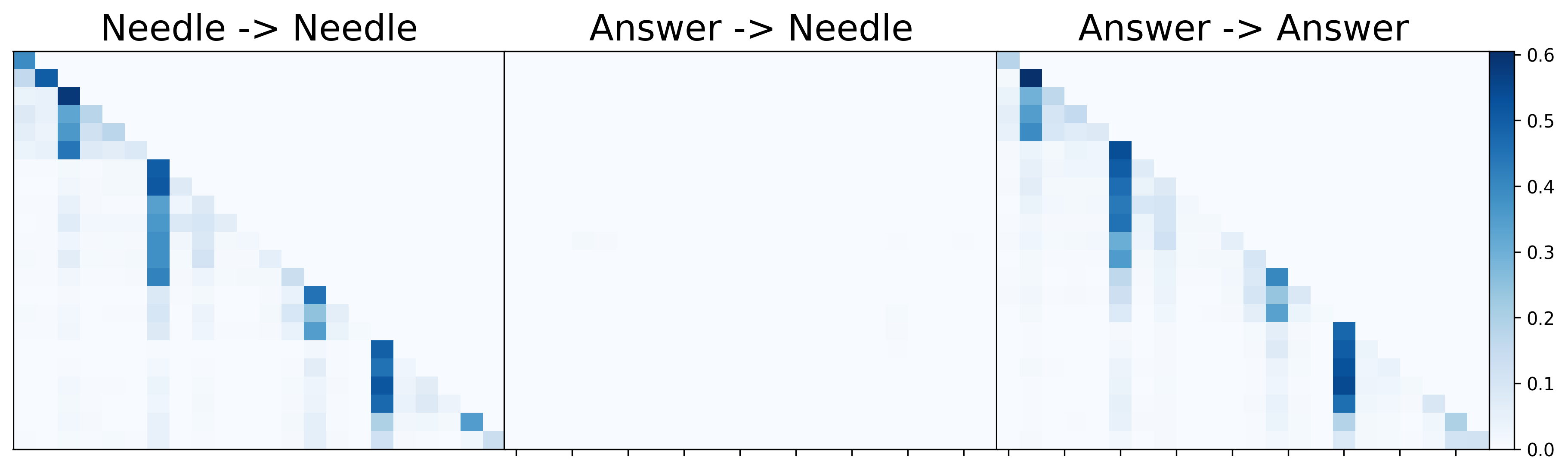} \\
    \includegraphics[width=0.95\linewidth] {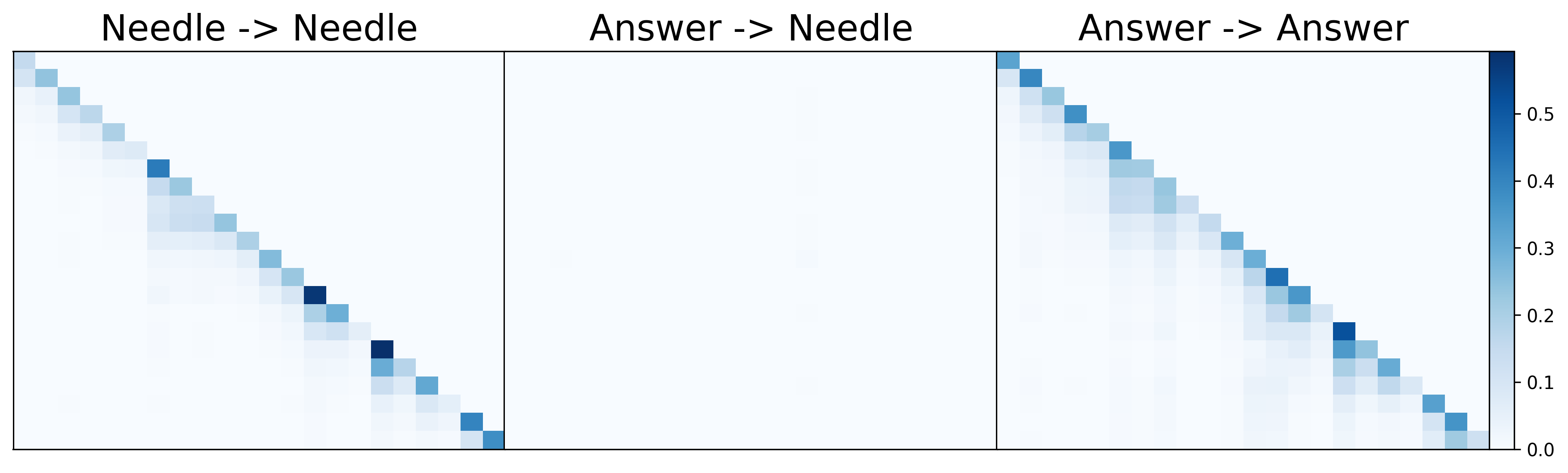} \\
    \caption{Visualization of attention scores from layer 15, head 5 across three models: Transformer (top), LoLCATs (middle), and LAWCAT (bottom). Each row presents three attention maps: needle-to-needle (left), answer-to-needle (center), and answer-to-answer (right)}
    \label{fig:viz:1}
\end{figure}

  Figure~\ref{fig:viz:1} illustrates an attention head primarily focused on local contextual information. A comparative analysis of the heatmaps reveals that while the Transformer, LoLCATs, and LAWCAT models exhibit broadly similar local attention patterns, notable differences emerge. The LoLCATs model displays a tendency for its attention to be diffused by more distant preceding tokens, potentially diluting its focus. In contrast, LAWCAT maintains a sharper concentration on the immediately relevant local context, particularly the current token, mirroring the behavior of the standard softmax Transformer more closely. This suggests that the integrated Conv1D layer in LAWCAT effectively strengthens local bias. By design, the convolutional operation aggregates information from adjacent tokens, allowing the model to consolidate the importance of the local neighborhood onto the current token's representation, thereby achieving a more focused attention distribution akin to softmax attention.

\begin{figure}[!htbp]
    \centering
    \includegraphics[width=0.95\linewidth] {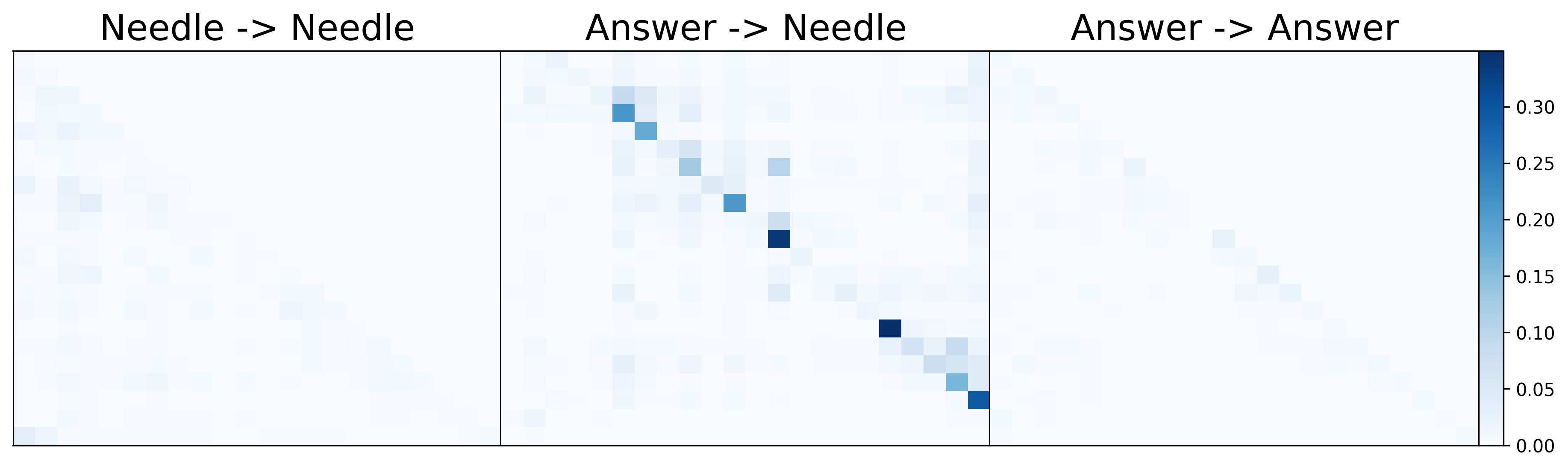} \\
    \includegraphics[width=0.95\linewidth] {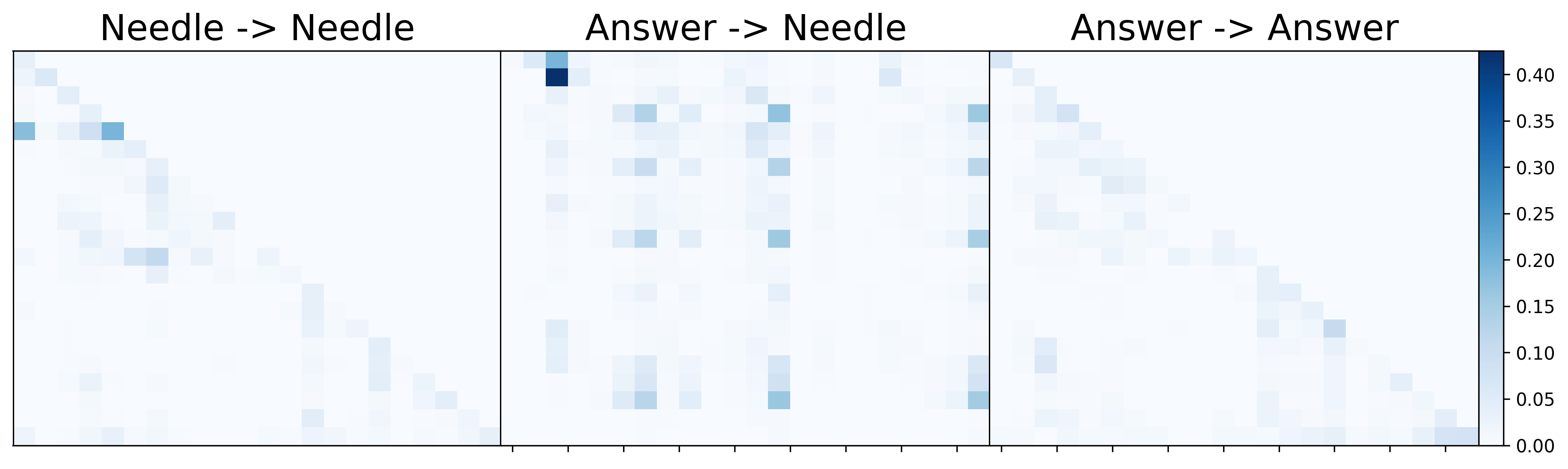} \\
    \includegraphics[width=0.95\linewidth] {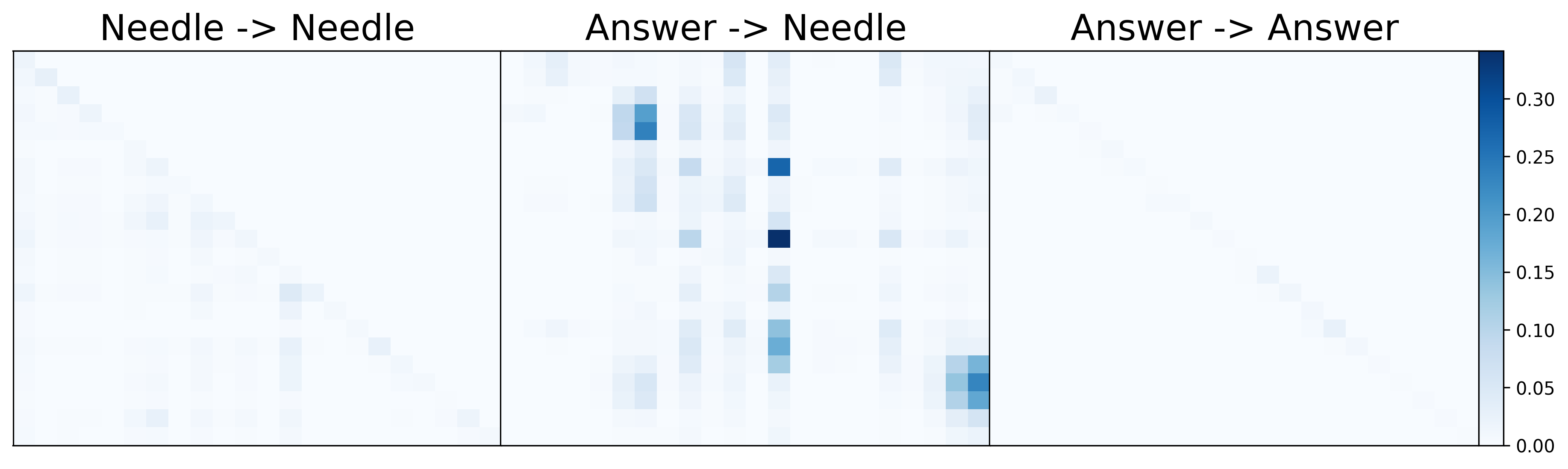} \\
    \caption{Visualization of attention scores from layer 15, head 22 across three models: Transformer (top), LoLCATs (middle), and LAWCAT (bottom). Each row presents three attention maps: needle-to-needle (left), answer-to-needle (center), and answer-to-answer (right)}
    \label{fig:viz:2}
\end{figure}

  Further insights are provided by Figure~\ref{fig:viz:2}, which depicts an attention head responsible for identifying the relationship between an answer and the "needle" (the target information to be retrieved). In this scenario, LoLCATs tends to allocate attention more broadly across the sequence, a characteristic that can hinder precise identification of the crucial answer-needle relationship. LAWCAT, however, mitigates this diffusion, exhibiting a more localized attention pattern that closely resembles the Transformer's focused engagement on the relevant segments. This closer alignment in attention scores between LAWCAT and the softmax Transformer not only correlates with LAWCAT's superior performance on retrieval tasks but also empirically validates the efficacy of the Conv1D layer in fostering a beneficial local inductive bias, enabling more precise attention allocation.

  \subsection{More Resutls on BABILong}
    \label{appendix:babilong}
\begin{figure}[t]
    \includegraphics[width=0.49\linewidth]{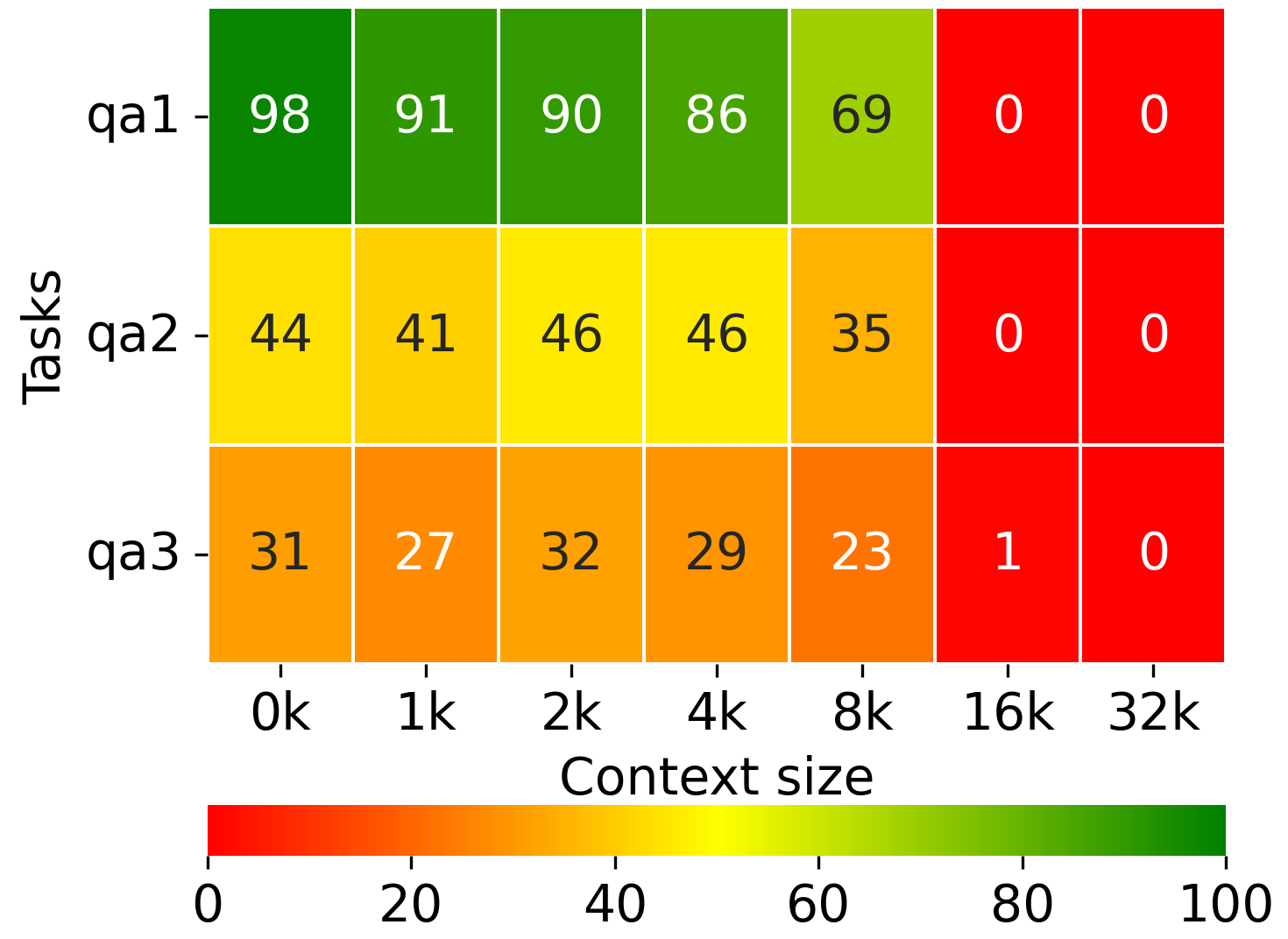}
    \hfill
    \includegraphics[width=0.49\linewidth]{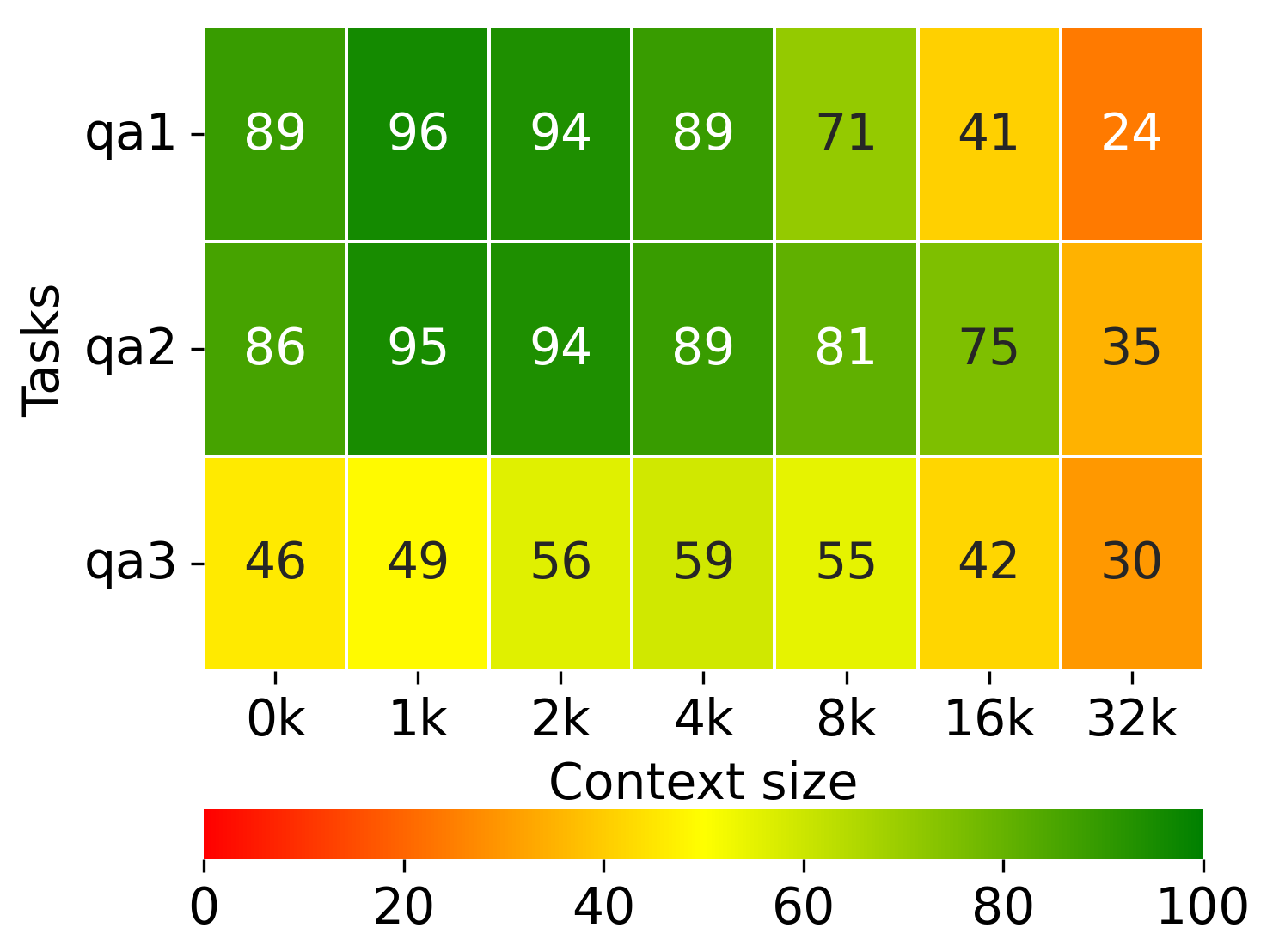}
    \caption{Comparison of accuracy on QA1\&QA2\&QA3 from BABILong benchmark between Llama 3 8B Instruct (left) and LAWCAT (right).}
    \label{fig:llama3-8b}
\end{figure}

\begin{figure}[t]
    \includegraphics[width=0.49\linewidth]{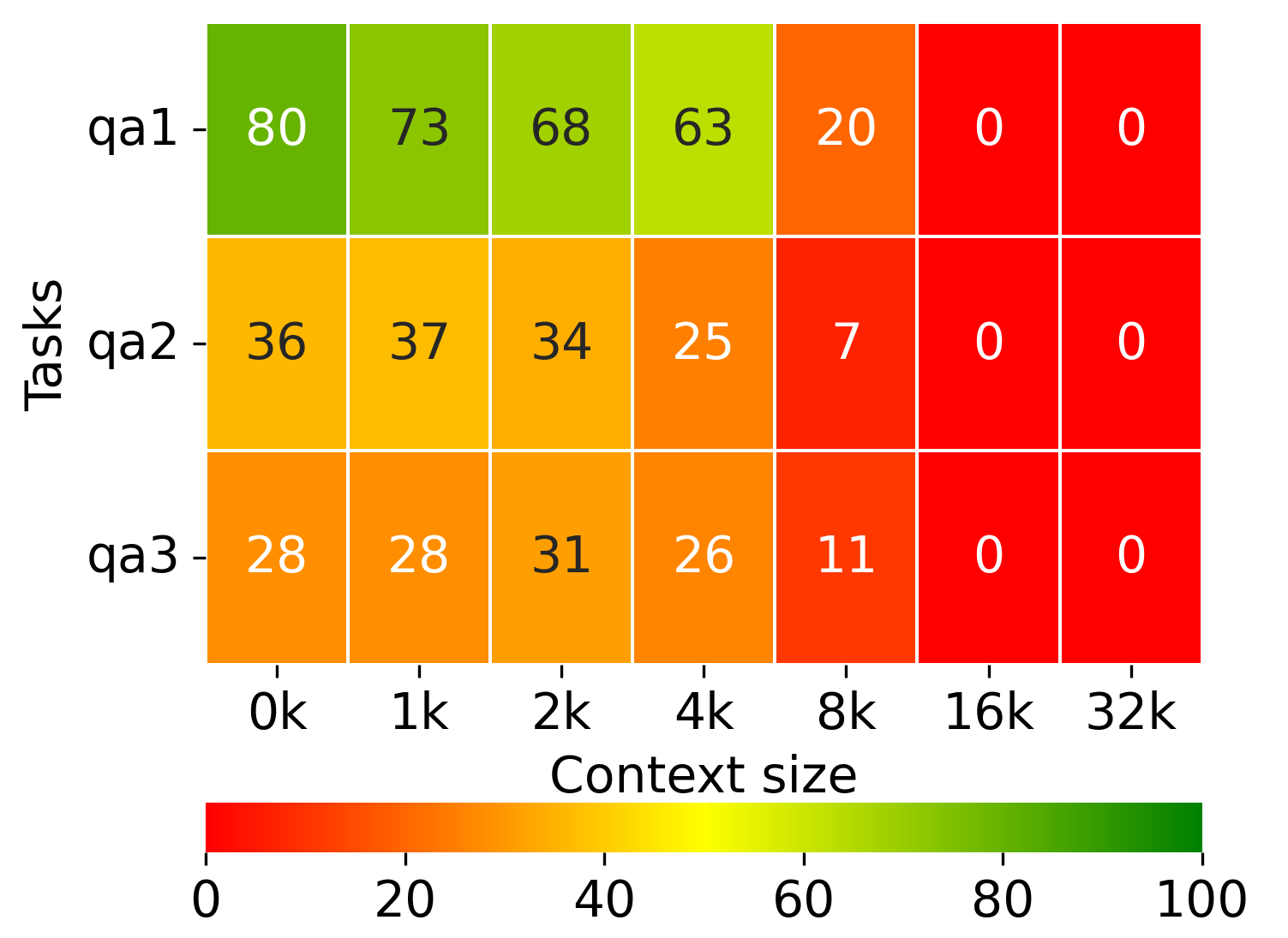}
    \hfill
    \includegraphics[width=0.49\linewidth]{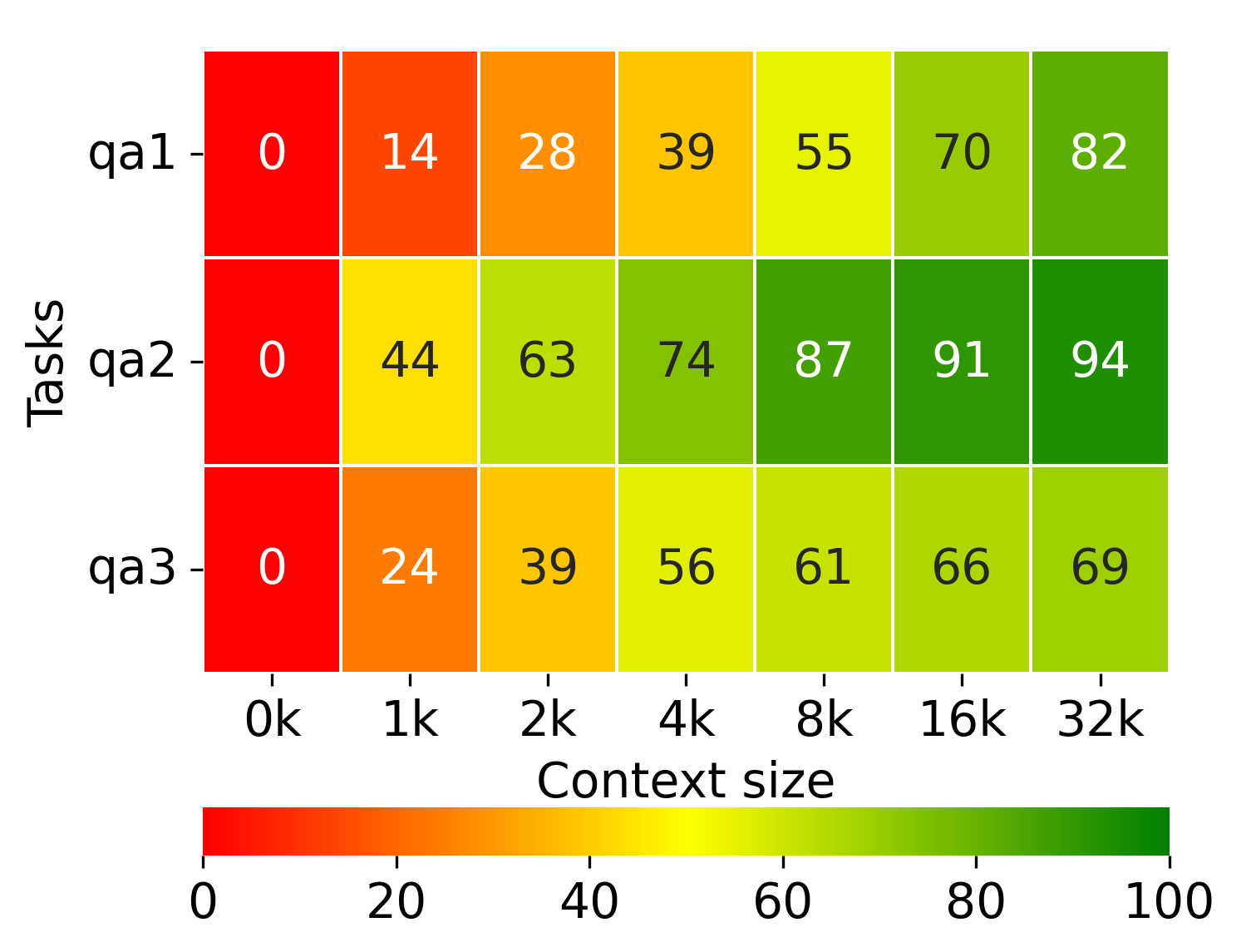}
    \caption{Comparison of accuracy on QA1\&QA2\&QA3 from BABILong benchmark between Mistral v0.3 7B (left) and LAWCAT (right).}
    \label{fig:mistral}
\end{figure}

We also added the rustles on BABILong of the Llama 3 8B Instruct and the LAWCAT variants, as shown in Fig.~\ref{fig:llama3-8b}, our LAWCAT variants consistently outperform the baselines, particularly at longer context lengths, mirroring the findings from experiments with the smaller LLaMA 3.2 1B model.
As shown in Fig.~\ref{fig:mistral}, compared to the original model, the distilled Mistral v0.3 7B model exhibits significantly improved performance on inputs longer than 8k tokens. However, it suffers from noticeable performance degradation on shorter context lengths, especially the 0k setting, which does not contain any "haystack" content. We hypothesize that this issue arises because Mistral v0.3 is a pretrained base model that is highly sensitive to prompt formatting, particularly after limited fine-tuning. We observed that, instead of producing incorrect answers, the model often outputs nothing for 0k inputs. 

\begin{table}[ht]
\centering
\begin{tabular}{lccccccc}
\toprule
Task & 0K & 1K & 2K & 4K & 8K & 16K & 32K \\
\midrule
\multicolumn{8}{c}{\textit{Pre-trained Transformer Model}} \\
QA2    & 44 & 52 & 47 & 35 & 14 &  5 &  2 \\
QA3    & 36 & 34 & 35 & 23 & 27 & 23 & 16 \\
\midrule
\multicolumn{8}{c}{\textit{LAWCAT}}     \\
QA2    & 94 & 97 & 95 & 90 & 82 & 32 &  8 \\
QA3    & 71 & 73 & 71 & 68 & 44 & 18 &  9 \\
\bottomrule
\end{tabular}
\caption{Results of Mistral v0.1 Instruct 7B on BABILong benchmark (QA2\&QA3)}
\label{tab:ins:mix1}
\end{table}

\begin{table}[ht]
\centering
\begin{tabular}{lccccccc}
\toprule
Task & 0K & 1K & 2K & 4K & 8K & 16K & 32K \\
\midrule
\multicolumn{8}{c}{\textit{Pre-trained Transformer Model}}    \\
QA2    & 48 & 41 & 31 & 19 &  3 &  2 &  1 \\
QA3    & 36 & 32 & 36 & 25 & 22 & 18 & 19 \\
\midrule
\multicolumn{8}{c}{\textit{LAWCAT}}    \\
QA2    & 87 & 92 & 95 & 96 & 96 & 72 & 43 \\
QA3    & 59 & 55 & 66 & 72 & 69 & 64 & 43 \\
\bottomrule
\end{tabular}
\caption{Results of Mistral v0.3 Instruct 7B on BABILong benchmark (QA2\&QA3)}
\label{tab:ins:mix3}
\end{table}

Therefore, we added the results of Mistral v0.1 Instruct 7B and Mistral v0.3 Instruct 7B models on the QA2 and QA3 tasks of the BABILong benchmark, as shown in Table~\ref{tab:ins:mix1} and Table~\ref{tab:ins:mix3} respectively.
As expected, applying LAWCAT to the models after instruction tuning will result in more stable performance, and the results of the Mistral models are consistent with those of the Llama model (although there are slight accuracy drops for the Mistral v0.1 7B Instruct LAWCAT variant model on QA3 when the context length is longer than 16k).

Overall, we believe these results further validate the effectiveness of our LAWCAT on various model architectures and tasks.

\end{document}